\documentclass[twocolumn]{article}

\usepackage[utf8]{inputenc}
\usepackage[T1]{fontenc}
\usepackage{graphicx}
\usepackage{amsmath,amssymb}
\usepackage[hidelinks, colorlinks=true, urlcolor=blue, linkcolor=black, citecolor=black]{hyperref}
\usepackage{authblk}
\usepackage{fancyhdr}
\usepackage{geometry}
\usepackage{caption}
\usepackage{float}
\usepackage{placeins}
\usepackage{dsfont}
\usepackage{booktabs}
\usepackage{algorithm}
\usepackage{algpseudocode}
\usepackage{array}
\usepackage{subcaption}
\usepackage{tikz}
\usepackage{atbegshi}
\AtBeginShipout{%
  \AtBeginShipoutUpperLeft{%
    \put(20,-700){\rotatebox{90}{\scriptsize arXiv:2501.12345v1 [cs.CV] \today}}
  }
}

\title{\textbf{Enhancing Cross-Domain SAR Oil Spill Segmentation\\
via Morphological Region Perturbation and Synthetic Label-to-SAR Generation}}

\author[1,2*]{Andre Juarez}
\author[1,2]{Luis Salsavilca}
\author[2]{Frida Coaquira}
\author[2]{Celso Gonzales}

\affil[1]{\small Círculo de Investigación de Máquinas de Aprendizaje (CIMA), Universidad Nacional Agraria La Molina, Lima 15024, Perú}
\affil[2]{\small Departamento de Estadística e Informática, Facultad de Economía y Planificación, Universidad Nacional Agraria La Molina, Lima 15024, Perú}

\date{} 

\begin{document}
\maketitle

\begin{figure*}[t] 
\centering 
\includegraphics[width=\textwidth]{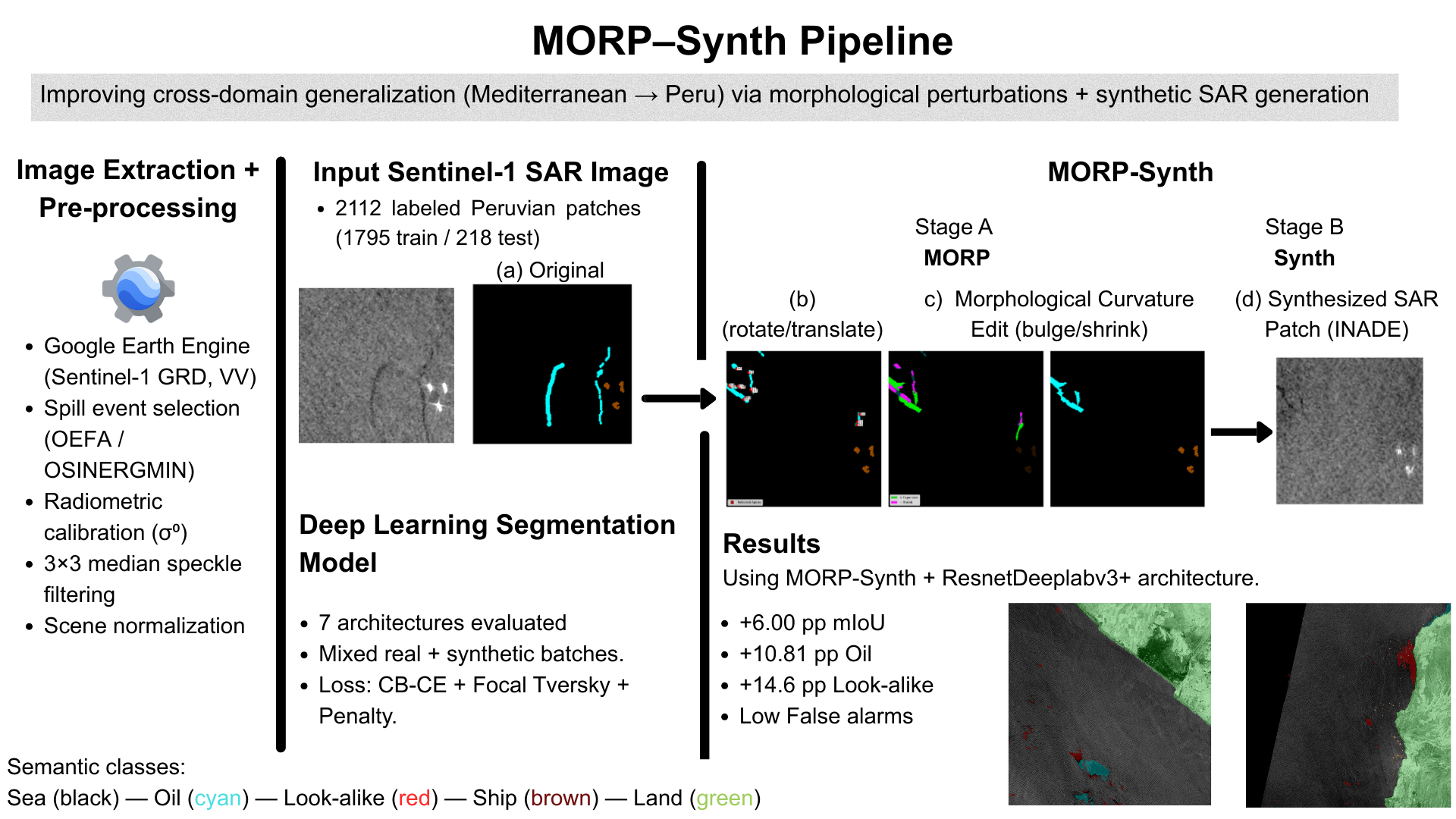} 
\caption*{\textbf{Graphical Abstract}. MORP--Synth framework overview.} 
\label{fig:graphical_abstract} 
\end{figure*}
\begin{abstract}
Deep learning models for SAR oil spill segmentation often fail to generalize across regions due to differences in sea-state, backscatter statistics, and slick morphology, a limitation that is particularly severe along the Peruvian coast where labeled Sentinel-1 data remain scarce. To address this problem, we propose \textbf{MORP--Synth}, a two-stage synthetic augmentation framework designed to improve transfer from Mediterranean to Peruvian conditions. Stage~A applies Morphological Region Perturbation, a curvature guided label space method that generates realistic geometric variations of oil and look-alike regions. Stage~B renders SAR-like textures from the edited masks using a conditional generative INADE model. We compile a Peruvian dataset of 2112 labeled 512$\times$512 patches from 40 Sentinel-1 scenes\\*
\mbox{(2014--2024)}, harmonized with the Mediterranean CleanSeaNet benchmark, and evaluate seven segmentation architectures. Models pretrained on Mediterranean data degrade from 67.8\% to 51.8\% mIoU on the Peruvian domain; MORP--Synth improves performance up to +6 mIoU and boosts minority-class IoU (+10.8 oil, +14.6 look-alike).
\end{abstract}

\footnotetext{*Correspondence: 20200396@lamolina.edu.pe (A.J.). 

Code Available at \\* \url{https://github.com/andrexandrex/MorpSynth.git}}

\vspace{4em}

\section{Introduction}
Marine oil spills are critical environmental hazards with long-lasting impacts on marine ecosystems, coastal livelihoods, and the economy. Rapid and accurate detection is essential for timely containment and mitigation measures. Among remote sensing techniques, Synthetic Aperture Radar (SAR), particularly Sentinel-1 imagery, has proven highly effective for oil spill monitoring due to its all-weather, day and night imaging capability~\cite{bianchiLargeScaleDetectionCategorization2020,al-ruzouqSensorsFeaturesMachine2020}. However, interpreting SAR images remains challenging: oil slicks appear as low-backscatter regions that are often indistinguishable from look-alike phenomena \cite{garcia-pinedaClassificationOilSpill2020,wanStudyVariationPatterns2024} such as low-wind areas, biogenic films, or naturally calm waters. This ambiguity demands automated segmentation techniques capable of reliably separating true oil spills from false positives.

Over the past decade, deep learning has transformed oil spill analysis, outperforming classical thresholding and texture based approaches. A major milestone was the release of a benchmark dataset containing 1,112 Sentinel-1 scenes with pixel-level labels for oil, look-alikes, sea, land, and ships \cite{krestenitisOilSpillIdentification2019,krestenitisEarlyIdentificationOil2019}. This benchmark enabled systematic evaluation of convolutional architectures such as U-Net \cite{ronnebergerUNetConvolutionalNetworks2015} and DeepLabV3\texttt{+} \cite{chenDeepLabSemanticImage2017}. Subsequent work introduced multi-scale contextual modeling and attention mechanisms to further reduce false positives \cite{zhuOilSpillContextual2022,satyanarayanaOilSpillSegmentation2023,dehghani-dehcheshmehOilSpillsDetection2023,buiOilSpillDetection2024}. These studies consistently demonstrate that modern deep neural networks surpass traditional SAR-based oil detection algorithms.

Despite these advances, geographic domain shift remains a major obstacle. Most segmentation models are trained on regional datasets primarily from European waters and struggle to generalize to distinct environmental regimes ~\cite{dehghani-dehcheshmehOilSpillsDetection2023,cuiEnhancedUnsupervisedDomain2025,kussulTransferLearningSinglepolarized2025}. The Southeast Pacific (Peruvian coast) illustrates this challenge: strong upwelling, the Humboldt Current, and unique wind conditions produce SAR backscatter textures and slick morphologies markedly different from those in the Mediterranean or North Atlantic (Table~\ref{tab:domain_differences}). This mismatch causes foreign-trained models to degrade when applied locally. Furthermore, annotated SAR datasets for South America remain scarce, limiting the applicability of state-of-the-art models. The 2022 Ventanilla/REPSOL spill underscored the urgent need for region specific monitoring tools \cite{mogollonREPSOLOilSpill2023}.

Another limitation of these models is the scarcity of representative training data in the target domain. Variations in sea surface roughness, oil properties, and incidence angles significantly impact model transferability. In practice, models pretrained on Mediterranean data often confuse Peruvian look-alikes with oil and fail to recognize slicks exhibiting different shapes or scattering signatures. While small-scale fine-tuning can mitigate this~\cite{Wang2018DeepDA,pratapFineArtFinetuning2025}, the limited amount of labeled Peruvian SAR data fails to capture the full variability required for robust generalization.

To overcome data scarcity, recent studies have explored generative augmentation. Approaches like diffusion-based models and conditional GANs~\cite{moonDiffusionbasedDataAugmentation2025,sunUtilizingDeepLearning2024,buiIMPROVINGACCURACYOIL2023,buiOilSpillDetection2024}  have been used to simulate SAR imagery. However, these methods typically operate in image space or lack explicit control over object morphology. They often fail to generate the specific, irregular slick geometries such as thin, fragmented trails driven by the Humboldt Current that are critical for improving generalization under severe domain shift.

\textbf{Our approach.} To address cross-domain generalization, we combine transfer learning with synthetic data augmentation. First, we curate a new SAR dataset of 40 oil spill events along the Peruvian coast (2014–2024), annotated in full alignment with the Mediterranean benchmark. Second, we introduce \textbf{MORP–Synth}, a two-stage augmentation pipeline tailored for data-scarce domains. Stage~A applies a Morphological Region Perturbation (MORP) algorithm that modifies oil and look-alike shapes via controlled geometric edits, shifting, rotating, and smoothly warping connected components (Figure~\ref{fig:morp_workflow}). Stage~B employs a employs a conditional Generative Adversarial Network (cGAN) based on Instance Adaptive De-Normalization (INADE)~\cite{Park2019SPADE,tan2021inade} to generate realistic SAR-like textures aligned with the edited labels (Figure~\ref{fig:inade_synth_pairs}). By conditioning feature normalization on instance-level statistics, the model achieves high-fidelity synthesis that preserves spatial coherence and label-image consistency. Synthetic samples are then mixed with real Peruvian patches using a controlled synthetic-to-real weighting strategy.

Additionally, we adopt a training strategy tailored for severe class imbalance, combining class balanced cross entropy, focal Tversky losses~\cite{Abraham2019FocalTversky,Salehi2017Tversky,abrahamNovelFocalTversky2018,cuiClassBalancedLossBased2019}, confusion-aware penalties, hard negative mining, and multi scale patch sampling to improve robustness in operational monitoring scenarios.

\textbf{Contributions.} The main contributions of this work are:
\begin{itemize}
    \item \textbf{Cross-domain benchmark and analysis:} A new Peruvian Sentinel-1 oil spill dataset harmonized with the Mediterranean benchmark.
    \item \textbf{MORP-Synth augmentation:} A geometry-aware synthetic augmentation pipeline that generates diverse slick shapes and realistic SAR textures, improving mIoU by up to +6 percentage points and significantly boosting minority-class performance.
    \item \textbf{Enhanced training under imbalance:} A practical training regimen integrating composite losses, confusion penalties, hard negative mining, and multi-scale sampling, achieving state-of-the-art segmentation performance in Peruvian waters.
\end{itemize}

This article is organized as follows. Section~2 (\textit{Materials and Methods}) describes the study areas, datasets, preprocessing, MORP-Synth design, and model training. Section~3 (\textit{Results}) shows quantitative evaluations and qualitative analyses. Section~4 (\textit{Discussion}) outlines methodological insights and operational implications. Section~5 (\textit{Conclusions}) summarizes the findings.

\section{Materials and Methods}

\subsection{Study Area and Datasets}

\paragraph{Mediterranean (Source) Dataset}
The source domain in this study corresponds to the Mediterranean Sea and is based on the publicly available dataset compiled by \cite{krestenitisOilSpillIdentification2019,krestenitisEarlyIdentificationOil2019}. This dataset contains a total of 1,112 Sentinel-1 SAR images in Ground Range Detected (GRD) format with vertical-vertical (VV) polarization. Each image is centered around known oil spill incidents reported in the CleanSeaNet (CSN) service operated by EMSA (European Maritime Safety Agency), and covers a spatial extent of 1250×650 pixels at a 10-meter ground resolution natively.

The dataset includes pixel-level segmentation masks distinguishing five semantic classes: open sea (0), oil spill (1), look-alike phenomena such as low wind or algae blooms (2), ships (3), and coastal land (4). Ground truth annotations were generated via visual inspection of SAR backscatter patterns in conjunction with metadata from CleanSeaNet alert records. This dataset provides a rich and balanced training ground for developing oil spill detection algorithms in a relatively stable marine environment.

\paragraph{Peruvian Coastal Dataset}

The Peruvian coastal dataset was developed to provide a representative target domain for oil spill detection under local environmental conditions. This region, located in the Southeast Pacific and influenced by the Humboldt Current, has oceanographic and meteorological characteristics that differ from those of the Mediterranean. These conditions make generalization harder, mainly because sea-surface roughness and slick dispersion behave differently along the Peruvian coast.

\begin{figure}[t]
    \centering
    \includegraphics[width=\columnwidth]
    {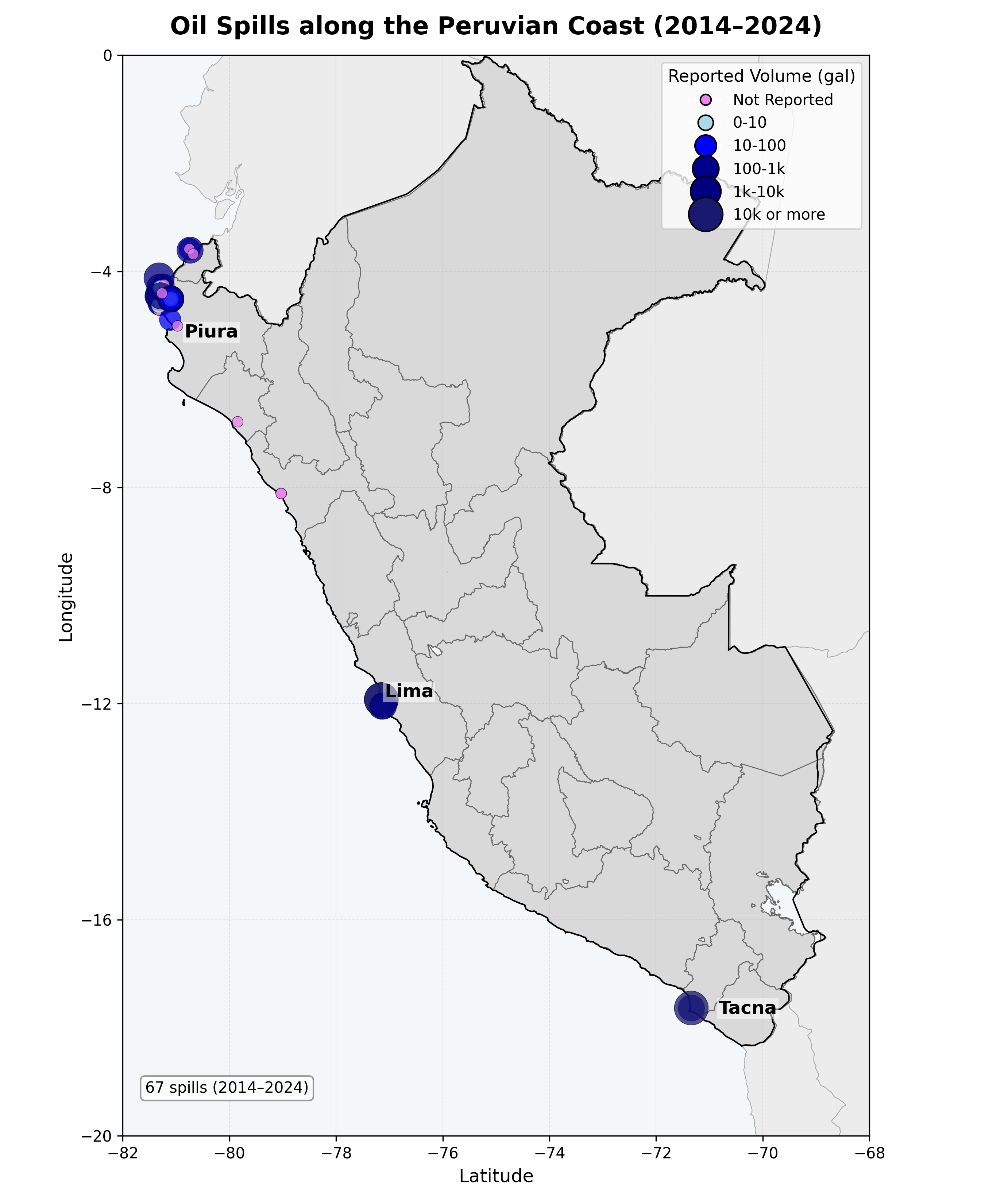}
    \caption{Spatial distribution and volume classification of 66 oil spills recorded along the Peruvian coast from 2012 to 2022.}
    \label{fig:peru_spills_map}
\end{figure}

The initial stage of dataset construction involved compiling 66 documented oil spill incidents from 2014 to 2022, based on official records from Organismo de Evaluación y Fiscalización Ambiental (OEFA) and Organismo Supervisor de la Inversión en Energía y Minería (OSINERGMIN). Figure~\ref{fig:peru_spills_map} visualizes the spatial distribution and estimated volumes of these incidents, which are mostly concentrated near Lima and the northern coast.

Figure ~\ref{fig:peru_spills_map} shows the spatial distribution of 67 reported oil spills (2012–2024) along the Peruvian coast, based on official records from OEFA and OSINERGMIN. Bubble size represents reported discharge volume. Most incidents cluster around Talara–Piura in the north, Callao–Lima in the center, and the Ilo-Tacna corridor in the south.

A total of 40 Sentinel-1 SAR images were ultimately selected from this set of events based on three criteria: (1) visual observability of the oil slick in the SAR image, (2) temporal proximity to the reported spill date (within 3–20 days, following \cite{mogollonREPSOLOilSpill2023}), and (3) availability and quality of satellite acquisition. These images span the period from 2014 to 2024 and were acquired from the Copernicus Sentinel-1 GRD mission with a size of 5569 x 5569 pixel resolution size.

Table~\ref{tab:peruvian_selection} in Appendix summarizes the 40 Sentinel-1 SAR scenes selected for the Peruvian coastal dataset, sorted chronologically by spill date. Each entry corresponds to a confirmed spill case, including its coordinates, SAR acquisition date, and delay in days between the incident and image capture.

\paragraph{Color Coding for Visualizations.}
\label{sec:study-area-datasets}
All segmentation maps in this work follow a consistent color scheme:
\textbf{oil (cyan)}, \textbf{look-alike (red)}, \textbf{land (green)}, \textbf{ship (brown)}, 
and \textbf{sea (black)}. Unless otherwise noted, this palette is used in all figures throughout 
the manuscript and Appendix.

\subsection{Preprocessing}
All Sentinel-1 GRD images were radiometrically calibrated to $\sigma^0$ and resampled to 10\,m/pixel. A $3\times3$ median filter was applied for mild speckle suppression; this choice preserves thin, elongated slick filaments that are easily oversmoothed by classical Lee-type filters\cite{leeSpeckleAnalysisSmoothing1981}. Only IW swath and VV polarization were retained. VV is known to be
more sensitive to rough sea-state modulation, which amplifies textural domain shift between calm Mediterranean scenes and the rougher Peruvian background.

\subsubsection{Domain Differences}
Despite relying on the same SAR sensor (Sentinel-1 GRD, VV polarization), the Mediterranean and Peruvian datasets exhibit distinct environmental and acquisition conditions that pose challenges for generalization and motivate the use of domain adaptation techniques (see Table~\ref{tab:domain_differences}).

These domain shifts affect the visual appearance of oil slicks and background patterns in SAR images, which can degrade the performance of models trained in one region when applied to another. Therefore, domain adaptation becomes essential to improve generalization to new environmental conditions.

\begin{table}[H]
\centering
\caption{Key differences between the Mediterranean and Peruvian SAR datasets.}
\label{tab:domain_differences}
\begin{tabular}{p{2.5cm} p{2.8cm} p{3.0cm}}
\toprule
\textbf{Aspect} & \textbf{Mediterranean Dataset} & \textbf{Peruvian Dataset} \\
\midrule
Sea conditions & Calmer semi-enclosed seas & Rougher open-sea; irregular slicks \\
\midrule 
Oceanographic forcing & Moderate seasonal winds & Strong Humboldt Current; upwelling \\
\midrule 
SAR background & Smooth; high contrast & Textured; risk of confusion with waves \\
\midrule 
Acquisition geometry & Mostly descending orbits & Mixed orbits; wider angle variability \\
\midrule 
Ground truth quality & CleanSeaNet masks & OEFA/OSINERGMIN reports \\
\bottomrule
\end{tabular}
\end{table}

\FloatBarrier
\begin{table*}[t]
\centering
\caption{Summary of deep learning segmentation families and representative models evaluated in this study.}
\label{tab:segmentation_families}

\begin{tabular}{p{3.3cm} p{7.2cm} p{5.0cm}}
\toprule
\textbf{Family} & \textbf{General Description} & \textbf{Representative Models} \\
\midrule
\textbf{CNN-based} &
Convolutional neural networks extract hierarchical spatial features using local 
receptive fields, convolutional weight sharing, and strong inductive biases. These
architectures offer stable performance and strong generalization in limited-data
settings. &
ResNet-UNet (R34)~\cite{ronnebergerUNet2015,heDeepResidual2015}, 
ResNet-UNet+ASPP~\cite{chenDeepLab2017}, 
DeepLabV3 (R34)~\cite{chenDeepLab2017}, 
EfficientNetV2-M + DeepLabV3~\cite{tan2021efficientnetv2, chen2017rethinking} \\

\textbf{Transformer-based} &
Transformer encoders leverage global self-attention to capture long-range 
dependencies and large-scale spatial context, improving recognition of extended
structures in SAR scenes. &
Swin-UNet~\cite{Cao2021SwinUnetUPA} \\

\textbf{CNN--Transformer Hybrids} &
Hybrid architectures combine CNN-based local feature extraction with Transformer 
modules for global reasoning, achieving strong boundary preservation together with 
holistic scene understanding. &
TransUNet~\cite{Chen2021TransUNet} \\
\bottomrule
\end{tabular}
\end{table*}

\subsection{Deep Learning Segmentation Models} 
We benchmark six architectures following encoder–decoder principles commonly used in
semantic segmentation. CNN-based models employ convolutional encoders and UNet- or
DeepLab-style decoders. Transformer-based models such as Swin-UNet replace the encoder
with hierarchical Swin Transformer blocks while retaining a UNet decoder. Hybrid
architectures such as TransUNet integrate CNN feature extractors with Transformer 
modules to capture both local structure and global context. EfficientNetV2 paired 
with a DeepLabV3 decoder remains a pure CNN pipeline, combining a compound-scaled 
EfficientNet backbone with an ASPP-based segmentation head for robust multi-scale 
aggregation.

\textbf{ResNet-UNet (R34).}
A UNet-style architecture combining a ResNet-34 encoder with a symmetric decoder linked by skip connections. 
The residual backbone stabilizes gradient flow and enhances semantic abstraction, while the UNet decoder preserves 
spatial detail yielding a well-balanced baseline for remote-sensing segmentation tasks~\cite{ronnebergerUNet2015,heDeepResidual2015}.

\textbf{ResNet-UNet + ASPP (R34+ASPP).}
A variant of ResNet-UNet where the bottleneck is replaced by an Atrous Spatial Pyramid Pooling (ASPP) module. 
Dilated convolutions enlarge the receptive field and improve capture of heterogeneous slick shapes and multi-scale 
structures, consistent with the DeepLab design~\cite{chenDeepLab2017}.

\textbf{DeepLabV3 (R34).}
A lightweight decoder-free model that aggregates features via ASPP and uses a shallow skip connection for boundary 
refinement. Its reduced computational cost makes it suitable for large-scale SAR inference and noisy sea-surface 
conditions~\cite{chenDeepLab2017}.

\textbf{EfficientNetV2-M + DeepLabV3.}
A fully convolutional encoder–decoder design pairing an EfficientNetV2 backbone a medium-capacity 
compound scaled CNN with a DeepLabV3 head. Progressive learning and fused-MBConv blocks improve speed accuracy 
efficiency, while ASPP provides robust multi-scale contextual reasoning~\cite{tan2021efficientnetv2,chen2017rethinking}.

\textbf{Swin-UNet.}
A pure Transformer-based model built upon the hierarchical Swin Transformer. Shifted-window self-attention captures 
long-range dependencies while maintaining local coherence, making it effective when oil spills or look-alikes span 
extended regions~\cite{Cao2021SwinUnetUPA}. It serves as our \emph{Transformer-only} benchmark.

\textbf{TransUNet.}
A hybrid architecture injecting CNN feature extraction into a ViT encoder to combine strong local texture modeling 
with global context reasoning. This fusion is especially effective for suppressing large look-alike structures while 
preserving fine boundaries~\cite{Chen2021TransUNet}. It complements Swin-UNet by illustrating the benefits of mixing 
CNNs and Transformers.
\FloatBarrier
\begin{figure*}[t]
    \centering

    \begin{tabular}{%
    >{\centering\arraybackslash}m{0.24\textwidth}%
    >{\centering\arraybackslash}m{0.22\textwidth}%
    >{\centering\arraybackslash}m{0.22\textwidth}%
    >{\centering\arraybackslash}m{0.22\textwidth}%
    }
    \textbf{(a) Original} &
    \textbf{(b) Rotate \& Place  + Apices Selection} &
    \textbf{(c) Curvature  Edits (+ / –)} &
    \textbf{(d) Final  Augmented Image}
    \end{tabular}

    \vspace{0.7em}

    \includegraphics[width=\textwidth]{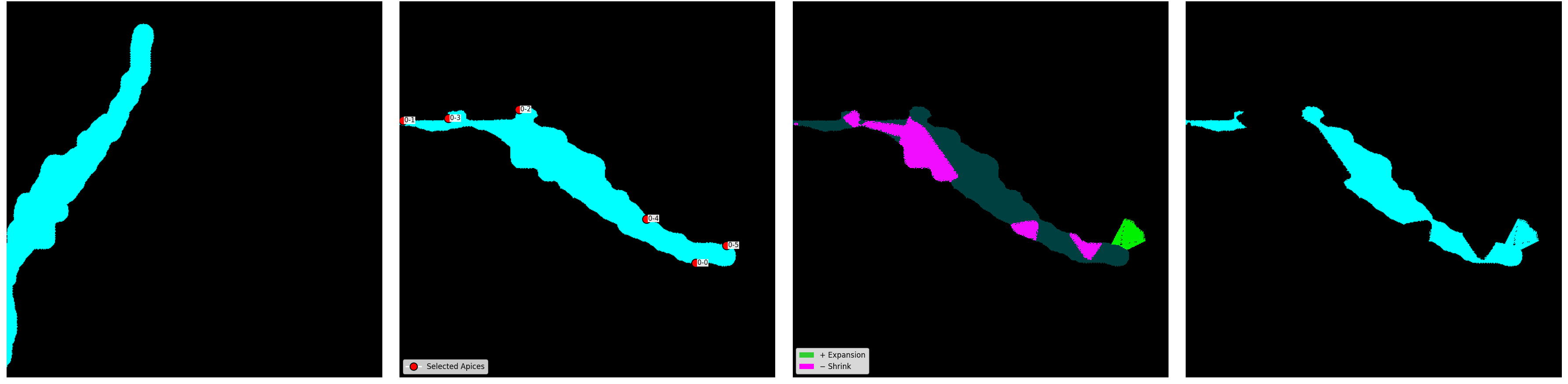}

    \vspace{1em}

    \includegraphics[width=\textwidth]{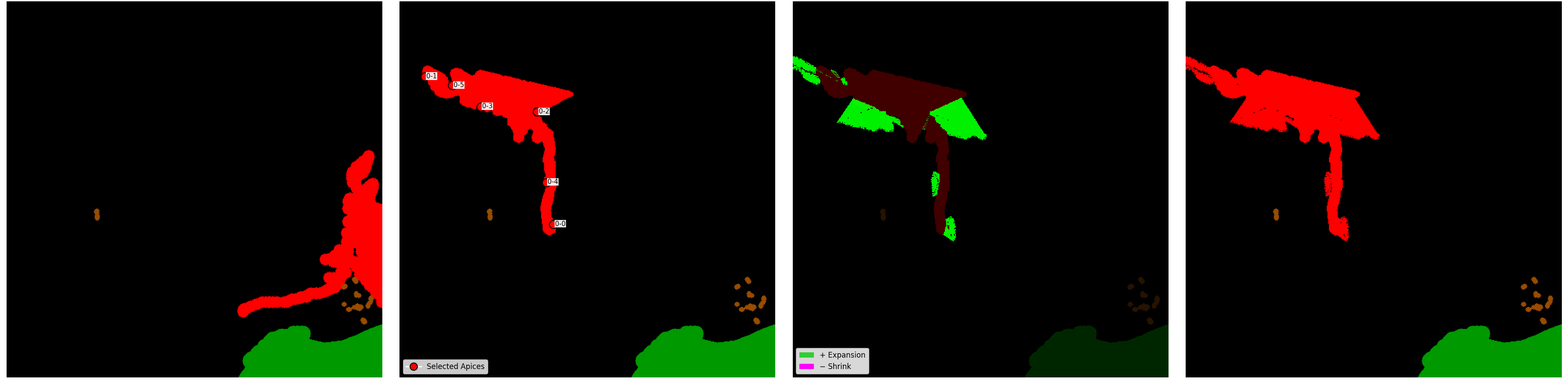}

    \caption{Illustration of the MORP augmentation workflow for two representative spill regions.
    Each row shows the four-stage process: 
    (a) original mask; 
    (b) rotated/translated region with detected apices; 
    (c) curvature-based edits (expansions in magenta; shrinkages in green); 
    (d) final augmented mask. Additional examples in Appendix.}
    \label{fig:morp_workflow}
\end{figure*}
\subsection{The MORP-Synth Pipeline}
\label{sec:morp-synth}

We propose \textbf{MORP-Synth}, a two-stage pipeline that augments scarce SAR oil-spill training data by
(1) perturbing object \emph{shape} directly in label space via our Morphological Region Perturbation (\textbf{MORP}), and
(2) rendering SAR-like appearance from the edited masks using an off-the-shelf conditional generator (INADE).
Our novelty lies in Stage~A; Stage~B is an applied component selected for efficiency and controllability.

\subsubsection{Stage A: Morphological Region Perturbation (MORP)}
\label{sec:morp-main}
Let $\mathcal{L}\in\{0,1,2,3,4\}^{H\times W}$ denote a semantic label map with classes
$0$ sea, $1$ oil, $2$ look-alike, $3$ ship, $4$ land. We define $\mathcal{C}=\{1,2\}$
as target classes for shape edits. MORP edits object geometry while preserving the
surrounding context. Its pipeline has three mathematically grounded components:
(i) \emph{flat-aware} rigid placement,
(ii) \emph{apex discovery} by smoothed outward curvature peaks, and
(iii) \emph{local apex edits} that mirror inward radial profiles to create outward
bulges (or the converse to carve wedges). The resulting augmented mask
$\mathcal{L}^\star$ is then used to condition the appearance model (Stage~B). The implementation code is provided in Appendix~\ref{alg:morp}.

\vspace{0.25em}
\paragraph{Connected components.}
For $c\in\mathcal{C}$, 

let $R\subset\{1,\dots,H\}\times\{1,\dots,W\}$ be a connected
component of $\mathds{1}[\mathcal{L}{=}c]$. We denote by $|R|$ its area in pixels
and by $\partial R$ its outer boundary (a simple, closed discrete curve).

\vspace{0.25em}
\paragraph{(1) Flat-Aware Region Placement.}
A region $R$ is rigidly transformed by a rotation $\theta\in[-\pi,\pi)$ and a
translation $\Delta=(\Delta x,\Delta y)$ with $\|\Delta\|_2\le S_{\max}$, producing
$T_{\theta,\Delta}(R)$. $S_{\max}$ (the \texttt{max\_shift} parameter) defines the maximum
translation distance. Placement is accepted only if it does not intersect
forbidden labels (\emph{e.g.}, land), expressed as a collision set $\Omega_{\text{forbid}}$.
We allow a whitelist $\Omega_{\text{allow}}$ for paste-over (e.g., sea, optionally ship).
Flat-aware perturbations: when $R$ exhibits long, near-linear boundary runs, small,
localized bulges may be added \emph{before} paste to avoid overly rectilinear silhouettes.

\paragraph{(2) Apex Detection via Smoothed Curvature. }
Parameterize $\partial R$ as a periodic sequence of points $P=\{(x_t,y_t)\}_{t=1}^N$
in counter-clockwise (CCW) order. We smooth coordinates with a circular
Savitzky–Golay filter (SG) ~\cite{savitzkySmoothingDifferentiationData1964}:
\begin{equation}
(\tilde x,\tilde y) \;=\; \mathrm{SG}\big((x,y);\, w,p,\mathrm{mode{=}}wrap\big),
\end{equation}

where $w$ is an odd window, $p$ a polynomial order ($p<w$).
Discrete derivatives use step $d_s\ge 1$:
\begin{equation}
\tilde x'=\nabla_{d_s}\tilde x,\quad \tilde y'=\nabla_{d_s}\tilde y,\quad
\tilde x''=\nabla_{d_s}\tilde x',\quad \tilde y''=\nabla_{d_s}\tilde y'.
\end{equation}
Signed curvature is discretized as:
\begin{equation}
\kappa_t \;=\; \frac{\tilde x'_t\,\tilde y''_t - \tilde y'_t\,\tilde x''_t}
{\big(\tilde x'^2_t+\tilde y'^2_t\big)^{3/2} + \varepsilon},\qquad \varepsilon>0.
\end{equation}
We use positive curvature $\kappa_t^+=\max(0,\kappa_t)$. To emphasize distal lobes,
we optionally apply a radial boost:
\begin{equation}
\kappa^{+}_t \;\leftarrow\; \kappa^{+}_t\Big(1+\rho\,\frac{r_t-\bar r}{\sigma_r+\varepsilon}\Big),\quad
r_t=\sqrt{(\tilde x_t-\bar x)^2 + (\tilde y_t-\bar y)^2},
\end{equation}
where $(\bar x,\bar y)$ is the centroid of $(\tilde x,\tilde y)$ and $\rho$ is a scalar hyperparameter controlling the 
boost strength.

Apex indices are the peaks of $\kappa^+$ exceeding a quantile threshold
$\tau=\mathrm{Quantile}(\{\kappa^{+}_t\},q)$ with minimum arc-distance $d$:
\begin{equation}
\mathcal{A}\;=\;\mathrm{Peaks}\!\big(\kappa^+;\,\mathrm{prom}>\tau,\ \mathrm{dist}\ge d\big).
\end{equation}
Apex coordinates are $A=\{(x_t,y_t): t\in\mathcal{A}\}$.
The detailed steps for this process are shown in Algorithm~\ref{alg:apices}.

\vspace{0.25em}
\paragraph{(3) Apex-based Perturbation via Mirrored Radial Profiles.}
To select at most $m$ well-spread apices, we cluster $A$ by $k$-means ($k=m$) and pick
in each cluster the point maximizing distance to the region centroid; denote the set by $A_m$.
Let $D$ be the Euclidean Distance Transform (DT) computed on the background pixels ($p \notin R$), denoted as $D=\mathrm{DT}(\mathds{1}[R^c])$.
At a chosen apex $a=(x_0,y_0)\in A_m$, an outward unit normal is:
\begin{equation}
\mathbf{n}(a) \;=\; \frac{\nabla D(a)}{\|\nabla D(a)\|_2+\varepsilon},
\end{equation}
with fallback $\mathbf{n}(a)\propto a - \mathrm{centroid}(R)$ if $\|\nabla D(a)\|$ is too small.
Consider a fan of $n_{\mathrm{rays}}$ unit directions
$\mathbf{u}_k=\mathcal{R}_{\phi_k}\,\mathbf{n}(a)$, where
$\phi_k$ are evenly spaced in $[-\alpha, +\alpha]$.
For each direction we measure the inward radial support length

\begin{equation}
d^{\mathrm{in}}_k \;=\; \max\{t\ge 0:\ a - t\,\mathbf{u}_k \in R\},
\end{equation}
and define an initial \emph{outward} target length:
\begin{equation}
d^{\mathrm{target}}_k \;=\; \lfloor s\cdot d^{\mathrm{in}}_k + \xi\rfloor,\quad
s\ge 1,\ \ \xi\sim\mathcal{U}\{0,1\}.
\end{equation}
This length is then capped using a "soft easing" rule: if $d^{\mathrm{target}}_k \ge R_{\text{grow}}$,
the final length is stochastically rescaled:
$d^{\mathrm{out}}_k = \lfloor R_{\text{grow}} \cdot (0.7 + 0.3 \cdot \mathcal{U}[0,1]) \rfloor$.
Otherwise, $d^{\mathrm{out}}_k = d^{\mathrm{target}}_k$. Here, $R_{\text{grow}}$ is the maximum
growth radius (\texttt{max\_radius}).
The polygon with vertices $\{a + d^{\mathrm{out}}_k\,\mathbf{u}_k\}_{k=1}^{n_{\mathrm{rays}}}$,
closed at $a$, defines a bulge mask $B(a)$.
In \emph{shrink} mode, we reverse the normal ($-\mathbf{n}$) and intersect the polygon with $R$
to form a wedge $S(a)\subset R$. Multi-apex composition performs a union of all bulges
and a union of all wedges, then writes $(R\cup B)\setminus S$ back into $\mathcal{L}^\star$.
The logic for a single apex edit is detailed in Algorithm~\ref{alg:apexedit}

\subsubsection{Stage B: Mask-to-SAR Synthesis (Synth)}
\label{sec:gen}

In the second stage, denoted \textbf{Synth}, we render SAR-like backscatter onto the edited semantic masks $\mathcal{L}^\star$ produced by MORP (Stage~A). We employ an INADE generator~\cite{tan2021inade}, an instance-adaptive extension of SPADE~\cite{parkSemanticImageSynthesis2019}. SPADE injects spatially aligned normalization parameters derived from the semantic mask, preventing the semantic layout from being washed out during generation, while INADE introduces instance-specific modulation to diversify textures across objects belonging to the same class. This is important for SAR oil-spill data, where \textit{oil} and
\textit{look-alike} regions can exhibit similar global darkness but distinct local
texture.

Unlike end-to-end diffusion models, INADE offers a favorable trade-off between
controllability and compute efficiency: it preserves mask-level semantics and allows
large batches of synthetic samples to be generated during training without
prohibitively slow sampling~\cite{dhariwalDiffusionModelsBeat2021,yangDiffusionModelsComprehensive2025}.
The generator and discriminator architectures used in Synth follow the standard
SPADE/INADE formulation, while training details (losses, hyperparameters, and
TTUR schedule) are provided in Sec.~\ref{inade_training}.

\FloatBarrier
\begin{figure*}[t]
    \centering

    \begin{tabular}{%
        >{\centering\arraybackslash}m{0.24\textwidth}%
        >{\centering\arraybackslash}m{0.22\textwidth}%
        >{\centering\arraybackslash}m{0.22\textwidth}%
        >{\centering\arraybackslash}m{0.22\textwidth}%
    }
    \textbf{(a) Real SAR} &
    \textbf{(b) Real Mask} &
    \textbf{(c) Synthesized Mask (MORP)} &
    \textbf{(d) Synthesized SAR (INADE)} \\
    \end{tabular}
    \vspace{0.6em}
    \includegraphics[width=0.24\textwidth]{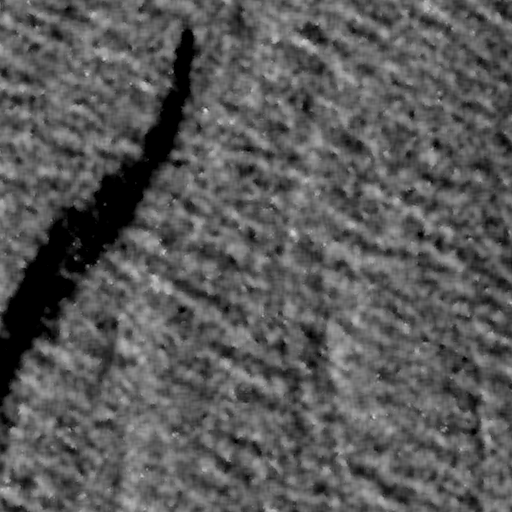}
    \includegraphics[width=0.24\textwidth]{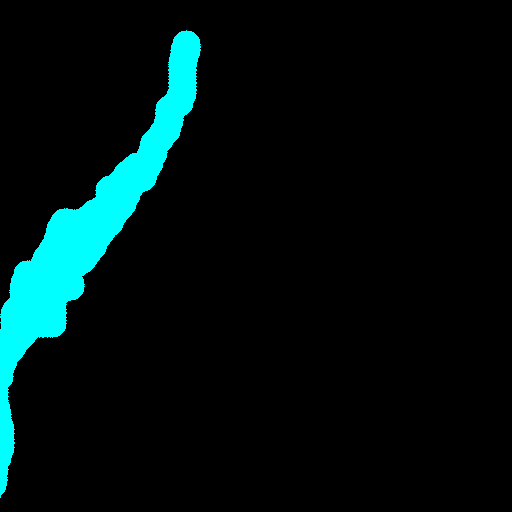}
    \includegraphics[width=0.24\textwidth]{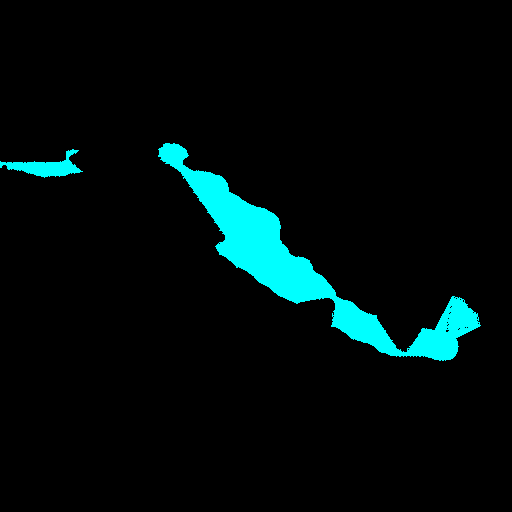}
    \includegraphics[width=0.24\textwidth]{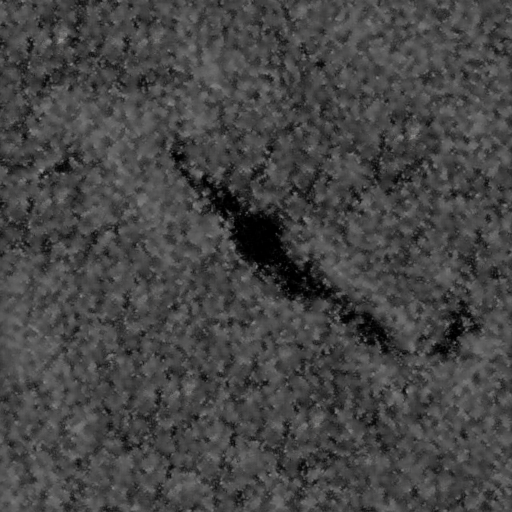}
    \\[0.9em]

    \includegraphics[width=0.24\textwidth]{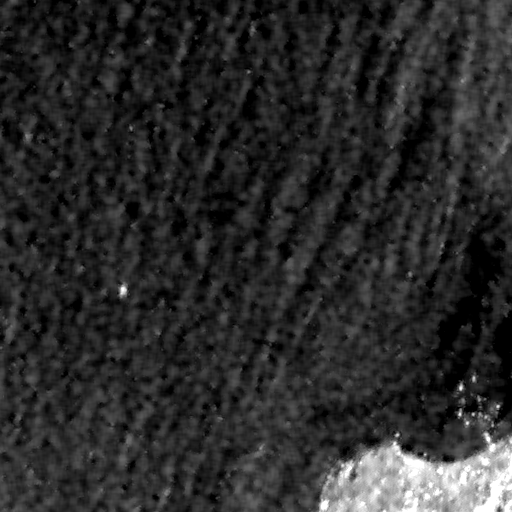}
    \includegraphics[width=0.24\textwidth]{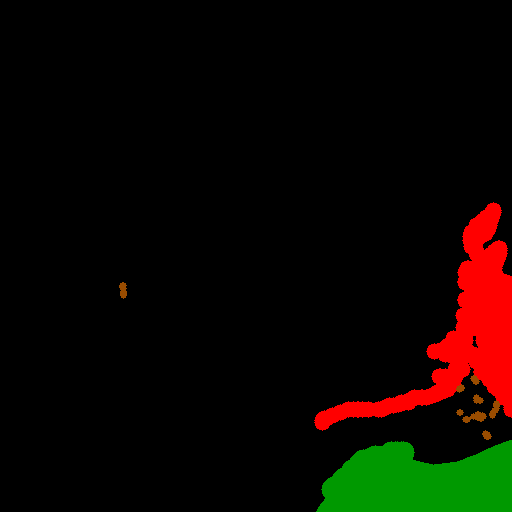}
    \includegraphics[width=0.24\textwidth]{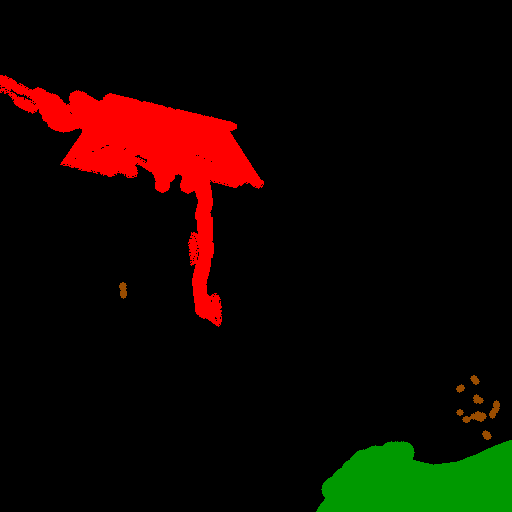}
    \includegraphics[width=0.24\textwidth]{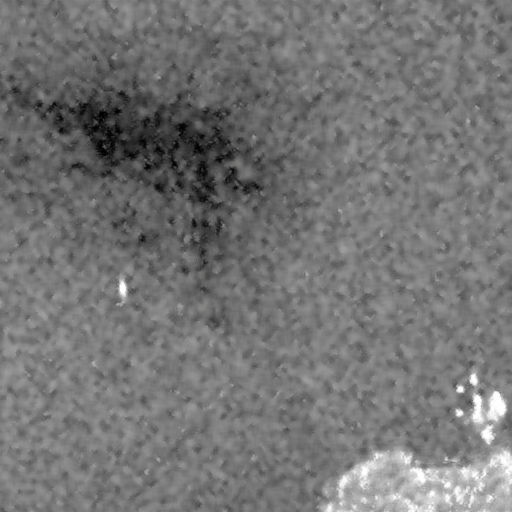}

    \caption{Comparison of real and synthesized samples in the Mask-to-SAR generation stage.
    Columns: (a) real Sentinel-1 SAR patches; 
    (b) ground-truth masks; 
    (c) augmented masks produced by \textbf{MORP}; 
    (d) synthetic SAR generated by \textbf{INADE}.}
    \label{fig:inade_synth_pairs}
\end{figure*}

\subsubsection{Putting it together}
Given a real mask $\mathcal{L}$, MORP produces an augmented shape $\mathcal{L}^\star$ (Stage~A), which conditions the Synth model (Stage~B) to yield a synthetic SAR patch $x^\star$.
We pair the synthetic samples $\{x^\star,\mathcal{L}^\star\}$ with real data $\{x,\mathcal{L}\}$ to train a segmentation model.
The complete transformation process is visualized in Figure~\ref{fig:morp_workflow} (MORP) and Figure~\ref{fig:inade_synth_pairs} (Synth).

\subsection{Training Details}
\label{sec:training}

All models were first trained from scratch on the Mediterranean dataset of~\cite{krestenitisOilSpillIdentification2019} 
and subsequently fine-tuned on the Peruvian SAR dataset (1,795 training / 218 testing patches). 
Scene-level splits were used to prevent spatial or temporal leakage between domains 
(See Appendix Table~\ref{tab:peruvian_selection}).  All models share identical preprocessing, augmentations, and loss formulation; 
differences in performance arise solely from backbone capacity and the inclusion of 
multi-scale and synthetic (MORP-Synth) data streams.

\paragraph{Patch extraction and normalization.}
Fixed-size $512{\times}512$ patches were sampled around annotated oil and look-alike regions 
using a sliding window and complemented by a limited number of background tiles 
(1.25:1 negative/positive ratio). 
Each patch was normalized by clipping backscatter values to the [0.5,97.5] percentile range per scene, 
preserving inter-scene contrast while mitigating illumination differences due to acquisition geometry or sea-state variability.  
Masks contained five semantic classes (\textit{sea, oil, look-alike, ship, land}),
and augmentations included horizontal and vertical flips, Gaussian noise, and Coarse Dropout.  
This procedure produced a radiometrically standardized collection of Peruvian oil-spill patches for subsequent fine-tuning.

\paragraph{Optimization.}
All networks were trained using the AdamW optimizer ($\beta_1{=}0.9$, $\beta_2{=}0.999$, weight decay $10^{-4}$) 
with a batch size of 32 (16 for heavier backbones). 
Mediterranean pretraining lasted 60~epochs at an initial learning rate of $1{\times}10^{-4}$, 
while Peruvian fine-tuning ran for 40~epochs at $2{\times}10^{-4}$, 
employing \textit{ReduceLROnPlateau} (factor~0.3, patience~5) and early stopping (patience~10). The best epoch was selected according to mean Intersection-Over-Union (mIoU) on the validation subset.  

\paragraph{Loss function.}
We minimize a composite objective that combines 
(i) class balanced cross entropy (CB--CE) to counter class imbalance ~\cite{cuiClassBalancedLossBased2019,heWeightingMethodsRare2021},
(ii) the Focal--Tversky loss~\cite{abrahamNovelFocalTversky2018,salehiTverskyLossFunction2017} to emphasize difficult foreground regions, and 
(iii) targeted confusion penalties to discourage systematic misclassifications 
between visually similar categories (look-alike$\!\to$oil and sea$\!\to$look-alike).  
The total real-data objective is given by:
\begin{equation}
\label{eq:loss_total}
\mathcal{L}_{\text{real}}
 = \lambda_{\mathrm{CE}}\,
   \mathcal{L}_{\mathrm{CB\mbox{-}CE}}
 + \lambda_{\mathrm{FTL}}\,
   \mathcal{L}_{\mathrm{Focal\mbox{-}Tversky}}
 + \sum_{(a,b)\in\mathcal{P}}
   \lambda_{a\!\to\!b}\,
   \mathcal{P}_{a\!\to\!b},
\end{equation}
where $(\lambda_{\mathrm{CE}},\lambda_{\mathrm{FTL}},\lambda_{2\to1},\lambda_{0\to2})
 =(0.3,\,0.7,\,0.40,\,0.30)$ and $\mathcal{P}=\{(2,1),(0,2)\}$.  
The first term, $\mathcal{L}_{\mathrm{CB\mbox{-}CE}}$, is the cross entropy loss weighted by:
\begin{equation}
w_c \propto \frac{1-\mu}{1-\mu^{n_c}}, 
\qquad 
\mu = \frac{\max_c n_c - 1}{\max_c n_c},
\end{equation}
where $n_c$ denotes the pixel count of class~$c$, and $\mu$ controls the effective volume.

The Focal--Tversky term is computed only over the foreground classes ($c>0$):
\begin{align}
\label{eq:focal_tversky}
\mathcal{L}_{\mathrm{Focal\mbox{-}Tversky}}
 &= \frac{1}{|\mathcal{C}_f|}
    \sum_{c\in\mathcal{C}_f}
    \big(1 - T_c\big)^{\gamma}, \\[0.3em]
T_c &= \frac{\mathrm{TP}_c}
           {\mathrm{TP}_c + \alpha\,\mathrm{FP}_c + \beta\,\mathrm{FN}_c + \varepsilon},
\end{align}
where $\mathrm{TP}_c, \mathrm{FP}_c, \mathrm{FN}_c$ are computed from 
softmax probabilities $p_{c,ij}$ and one-hot labels $y_{c,ij}$ as usual.  
We set $(\alpha,\beta,\gamma)=(0.65,0.35,1.33)$ to penalize false positives more strongly, 
with $\gamma>1$ focusing learning on low-Tversky regions.  
Finally, each confusion penalty term takes the form:
\begin{equation}
\label{eq:confusion_penalty}
\mathcal{P}_{a\to b}
 = \mathbb{E}\!\left[(p_b)^{\gamma_{\!p}} \,\middle|\, y=a\right],
\qquad
\gamma_{\!p}=2.0,
\end{equation}
averaging the model’s confidence for the wrong class~$b$ 
over all pixels whose true label is~$a$.  
This term acts as a soft constraint to reduce confident false alarms between 
semantically related categories.

\paragraph{Hard negatives and multi-scale augmentation.}
To increase robustness against false positives, we incorporated 210 hard-negative patches mined from the Peruvian training scenes \cite{liHardNegativesMining2025,chenDeepLabSemanticImage2017}.  
These correspond to background regions where a previously fine-tuned model produced confident spill detections 
(overlap~$\ge80\%$ with sea and area~$\ge0.5\%$).  
They were integrated via a weighted sampler contributing $\sim$10\% of real batches.  
In addition, 959 multi-scale patches (rescaled 1024–2048~pixel windows, resized to $512{\times}512$) 
were added to expand spatial context around slicks and coastal structures, 
representing roughly 20–25\% of effective real batches.  

\paragraph{Synthetic label-to-SAR augmentation (MORP-Synth).}
Synthetic samples were generated using the \textbf{MORP-Synth} pipeline, which integrates
(i) label-space perturbations via \textbf{MORP} (Stage A) and
(ii) conditional appearance synthesis via the \textbf{Synth} (INADE) generator trained on the Mediterranean domain~\cite{tanDiverseSemanticImage2021}.
The parameter \texttt{morph} controls the fraction of label maps undergoing MORP perturbation before synthesis,
with \texttt{morph00}, \texttt{morph50}, and \texttt{morph100} corresponding to
0\%, 50\%, and 100\% edited masks, respectively.
Synthetic batches were interleaved with real ones at a $2{:}1$ ratio, and the training objective extended Eq.~\ref{eq:loss_total} as:
\begin{equation}
\label{eq:loss_synth}
\mathcal{L}_{\text{total}}
 = \mathcal{L}_{\text{real}} + \lambda_{\text{synth}} \, \mathcal{L}_{\text{synth}},
\end{equation}
where $\lambda_{\text{synth}}\in\{25,50,100,150,200\}$
scales the contribution of synthetic gradients without changing sample counts.
Small $\lambda_{\text{synth}}$ values limit exposure to synthetic features,
while larger ones emphasize synthetic regularization.

\paragraph{Synth model training (Stage B).}
\label{inade_training}
The INADE generators were trained for 60~epochs
using Adam with TTUR-style learning rates~\cite{heuselGANsTrainedTwo2018}
($D{:}G{=}1{:}1$, $\lambda_\text{feat}{=}10$, $\lambda_\text{VGG}{=}10$, $\text{lr}{=}5{\times}10^{-5}$)
and a spectrally normalized discriminator with hinge loss~\cite{luoLearningSmoothHinge2021}.
The best checkpoint (epoch~42, FID~$\approx$~61) produced visually coherent SAR-like backscatter
and was adopted for subsequent mixed-domain training.

\subsubsection{Evaluation Metrics}
\label{sec:evaluation}

\paragraph{Mean Intersection-over-Union (mIoU)}  
We evaluate segmentation performance using the mean Intersection over Union (mIoU) over all \(C=5\) semantic classes. For each class \(c \in \{0,\dots,4\}\), the IoU is defined as:
\begin{equation}
\mathrm{IoU}_c = \frac{\mathrm{TP}_c}{\mathrm{TP}_c + \mathrm{FP}_c + \mathrm{FN}_c},
\end{equation}
where \(\mathrm{TP}_c\), \(\mathrm{FP}_c\), and \(\mathrm{FN}_c\) are the number of true positives, false positives, and false negatives for class \(c\), respectively.

\section{Results}

\subsection{Quantitative Evaluation}

\paragraph{Baseline (Source-only)}

After pretraining on the Mediterranean source dataset ~\cite{krestenitisEarlyIdentificationOil2019,krestenitisOilSpillIdentification2019}, all seven architectures exhibited stable convergence within 60 epochs, reaching validation mIoUs between 0.75 and 0.85 with no evidence of overfitting. The learned features from this stage served as initialization for transfer to the Peruvian target domain, where models were either fine-tuned from their Krestinin checkpoints (dashed lines) or retrained from standard ImageNet-pretrained weights (dotted lines). Fine-tuned runs started from the validation performance achieved on Krestinin and consistently converged faster and to higher mIoU and lower loss than ImageNet-initialized counterparts, confirming effective feature reuse across marine environments despite differences in sea-state and incidence-angle statistics. 

Table~\ref{tab:krestinin_test} summarizes the performance of all evaluated architectures on the unseen Krestinin test set (\(n{=}569\) tiles of \(512\times512\)~pixels). 
Across all baselines, the newly introduced \textit{Swin-Tiny UNet} achieved the highest mean IoU (68.66\%), followed closely by the \textit{ResNet-34 DeepLabV3\texttt{+}} (67.82\%) and \textit{ResNet-34 UNet+ASPP} (66.05\%). 
The transformer-based Swin backbone offered improved boundary delineation and spatial consistency, particularly on the \textit{oil spill} (51.80\%) and \textit{look-alike} (53.76\%) classes, where contextual attention across wide receptive fields reduces confusion with textured marine backgrounds~\cite{Cao2021SwinUnetUPA,chenTransUNetTransformersMake2021}.

\begin{figure*}[H]
\centering
\includegraphics[width=\textwidth]{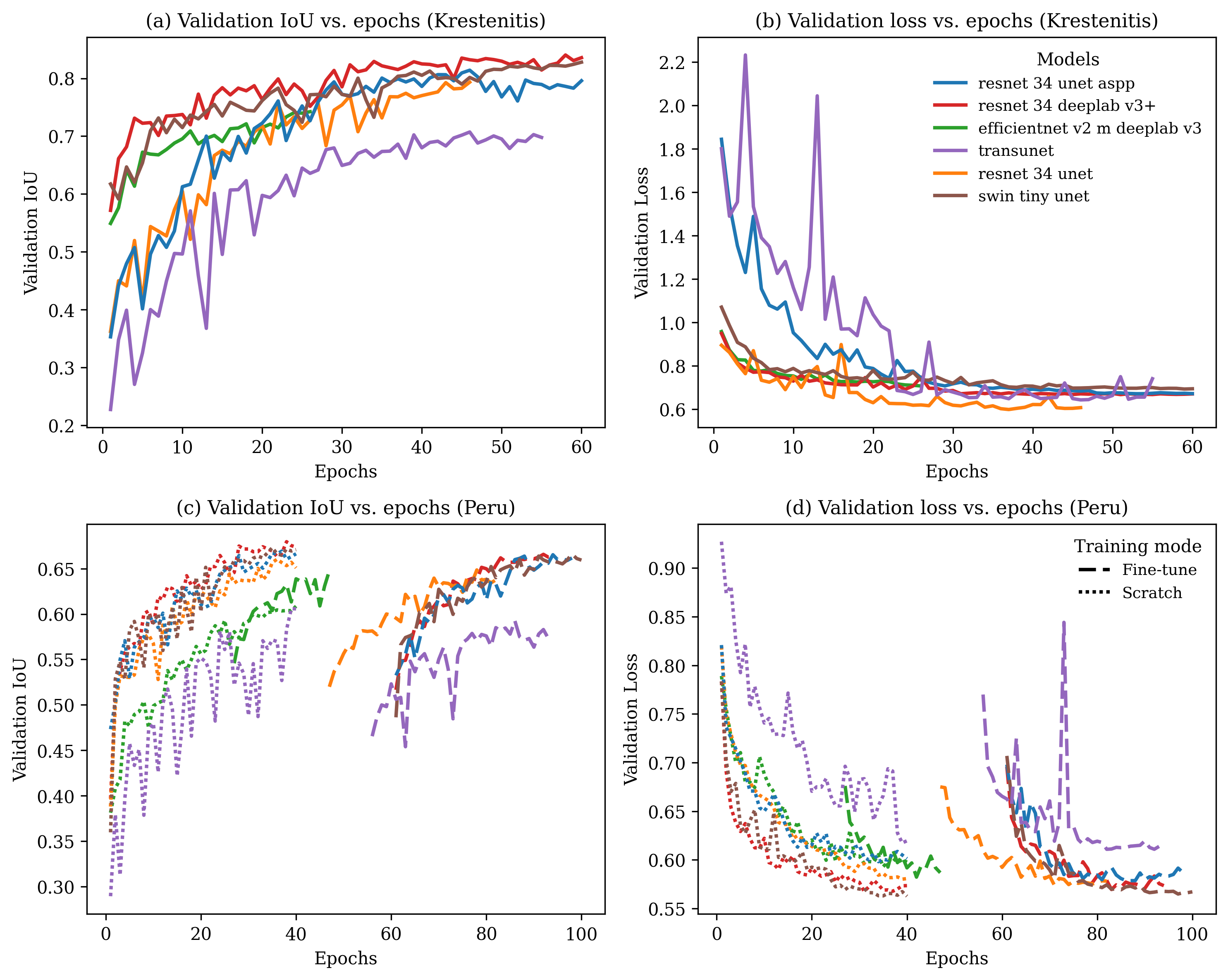}
\caption{Training curves on the Krestinin source domain (top row) and Peruvian target domain (bottom row).
(a) Validation mIoU on Krestinin. 
(b) Validation loss on Krestinin. 
(c) Validation mIoU on Peru. 
(d) Validation loss on Peru.}
\label{fig:training_krestinin}
\end{figure*}

Among convolutional models, integrating multi-scale contextual modules such as ASPP or DeepLab’s atrous encoders enhanced discrimination of elongated and low-contrast spill regions, confirming prior observations for SAR segmentation~\cite{charngDeepLearningSegmentation2020,dehghani-dehcheshmehOilSpillsDetection2023}. 
While \textit{TransUNet} reached 62.93\% mIoU, it underperformed relative to purely convolutional and Swin-based models, likely due to its heavier hybrid encoder and limited training data for learning long-range dependencies. 
The \textit{EfficientNetV2-M DeepLabV3} yielded comparable overall accuracy (62.09\%) but weaker results for the \textit{ship} class (23.24\%), consistent with the over-smoothing effect of compound scaling in compact maritime targets. 
Overall, the \textit{Swin-Tiny UNet} emerged as the most balanced and data-efficient architecture, delivering top mIoU and per-class consistency across both dominant (\textit{sea, land}) and minority (\textit{oil, look-alike, ship}) categories. 
Given its strong baseline performance, it was subsequently used to examine the generalization limits of the proposed MORP-Synth augmentation framework under a higher-capacity setting (See Table \ref{tab:swin_unet_tiny_synth_morph_scale_multi} .

\begin{table}[htbp]
\centering
\caption{Model performance on the unseen Krestinin test set (\(n = 569\) tiles). Results are reported as per-class IoU and mean IoU (\%).}

\label{tab:krestinin_test}
\FloatBarrier
\resizebox{\columnwidth}{!}{%
  \begin{tabular}{lcccccc}
    \toprule
    \textbf{Model}                  & \textbf{mIoU} & \textbf{Sea} & \textbf{Oil} & \textbf{Look-alike} & \textbf{Ship} & \textbf{Land} \\
    \midrule
    \textit{TransUNet}              & 62.93         & 93.16        & 44.91        & 49.22               & 43.16         & 84.19         \\
    \textit{ResNet-34 UNet}         & 65.82         & 94.60        & 50.77        & 43.74               & 45.50         & 94.50         \\
    \textit{ResNet-34 UNet+ASPP}    & 66.05         & 94.45        & 50.64        & 52.49               & 39.59         & 93.08         \\
    \textit{ResNet-34 DLV3\texttt{+}} & 67.82 & \textbf{94.66} & 50.40        & 51.00               & \textbf{47.35} & 95.68         \\
    \textit{EfficientNet DLV3}    & 62.09         & 94.22        & 47.34        & \textbf{54.28}      & 23.24         & 91.36         \\
    \textit{Swin-Tiny UNet}         & \textbf{68.66} & 95.22        & \textbf{51.80} & 53.76               & 46.42         & \textbf{96.09} \\
    \bottomrule
  \end{tabular}%
}
\end{table}

\paragraph{Cross-domain Testing}
Modern segmentation \\* pipelines increasingly rely on large pre-trained models, yet direct deployment to out-of-domain imagery often leads to degraded performance due to domain shift~\cite{pratapFineArtFinetuning2025}. To assess cross-domain transferability, each architecture was fine-tuned on the Peruvian training split for 40~epochs using the composite loss defined in Eq.~\ref{eq:loss_total}. As summarized in Table~\ref{tab:peru_adaptation}, fine-tuning improved mean IoU for most architectures while preserving or enhancing class-wise balance.

Among convolutional networks, \textit{ResNet-34 UNet} achieved the strongest recovery, improving mean IoU by +4.9~pp and showing the largest per-class gains on \textit{oil} (+6.9~pp) and \textit{look-alike} (+9.3~pp), indicating better boundary delineation and reduced confusion between low-contrast slicks and sea background. \textit{TransUNet} and \textit{EfficientNetV2-M~+~DeepLabV3} also benefited from adaptation (+2.9~pp and +2.3~pp, respectively), reflecting partial reuse of spectral–textural priors learned in the source domain. 

\textit{ResNet 34 DeepLabV3} remained stable (+0.3~pp) while slightly improving ship detection (+4.7~pp).  In contrast, \textit{UNet+ASPP} exhibited a small decrease (--2.1~pp), mainly from degraded segmentation of the \textit{look-alike} class (--14.3~pp), consistent with ASPP’s known sensitivity to heterogeneous marine textures. This decline suggests that the strong contextual filters optimized on the Mediterranean domain over-specialized to its smoother sea clutter, leading to feature drift when exposed to the higher speckle and coastline variability of Peruvian SAR imagery.

The transformer-based \textit{Swin-Tiny UNet}—already the strongest source-only model showed marginal overall improvement (+0.1~pp) yet achieved the highest adapted IoUs for \textit{oil} (52.42\%). This limited gain suggests that global attention priors in Swin architectures already capture robust scene context, leaving less room for improvement from local fine-tuning alone. Overall, these results confirm that modest domain-specific retraining effectively bridges cross-domain gaps for convolutional backbones, whereas transformer-based models may require more sophisticated augmentation or loss weighting to further exploit their representation capacity.

\begin{table}[htbp]
\caption{Per-class IoU (\%) on the Peruvian test set before and after full fine-tuning. “Source-only” refers to models trained solely on Krestinin data; “Adapted” indicates full fine-tuning on Peruvian data. Best results per class in \textbf{bold}.}
\centering
\resizebox{\columnwidth}{!}{%
\begin{tabular}{lccccccc}
\toprule
\textbf{Model} & \textbf{Setup} & \textbf{mIoU} & \textbf{Sea Surface} & \textbf{Oil Spill} & \textbf{Look-alike} & \textbf{Ship} & \textbf{Land} \\
\midrule
\textit{ResNet-34 UNet} & Source-only & 51.80 & 91.28 & 37.50 & 21.16 & 16.18 & 92.90 \\
                        & Adapted     & \textbf{56.70} & 93.47 & 44.38 & 30.48 & 21.94 & 93.23 \\
\textit{ResNet-34 UNet+ASPP} & Source-only & 56.65 & \textbf{94.69} & 43.71 & 35.19 & 17.52 & 92.14 \\
                             & Adapted     & 54.57 & 94.64 & 40.89 & 20.89 & 22.93 & \textbf{93.51} \\
\textit{ResNet-34 DLV3} & Source-only & 53.79 & 93.66 & 38.16 & 25.86 & 17.92 & 93.34 \\
                              & Adapted     & 54.07 & 94.24 & 38.56 & 23.37 & \textbf{22.64} & 91.52 \\
\textit{TransUNet}            & Source-only & 51.74 & 93.34 & 39.94 & 23.11 & 13.36 & 88.94 \\
                              & Adapted     & 54.64 & 92.94 & \textbf{46.00} & 29.77 & 15.40 & 89.09 \\
\textit{EfficientNet DLV3} & Source-only & 49.25 & 91.76 & 35.00 & 23.59 & 5.81 & 90.09 \\
                                    & Adapted     & 51.51 & 93.99 & 36.48 & 23.10 & 11.08 & 92.87 \\
\textit{Swin-Tiny UNet}      & Source-only & 57.93 & 92.99 & 50.49 & 32.05 & 20.17 & 93.95 \\
                             & Adapted     & \textbf{58.03} & 93.07 & 52.42 & \textbf{31.97} & 20.05 & 92.62 \\
\bottomrule
\end{tabular}
}
\label{tab:peru_adaptation}
\end{table}

\paragraph{Synthetic label-to-SAR augmentation (MORP-Synth) with scale weighting}
\label{sec:morp_synth_results}

We assess the quantitative impact of synthetic augmentation using \textbf{ResNet-34 DeepLabV3\texttt{+}} as the primary analysis backbone (Tables~\ref{tab:krestinin_test},~\ref{tab:peru_adaptation}). Synthetic sets were generated with the INADE checkpoint described in Sec.~\ref{inade_training}. Three pool sizes were tested (\(N_{\text{synth}}\in\{902,1804,3608\}\)), with real:synthetic batches interleaved at \(2{:}1\) and synthetic gradients scaled by \(\lambda_{\text{synth}}\in\{25,50,100,150,200\}\).
For the base pool (\(N_{\text{synth}}{=}902\), matching the available positive seed patches), we compared four edit regimes: \textbf{nomove} (no geometric change), \textbf{m00} (movement only), \textbf{m50} (movement+MORP on 50\% of masks), and \textbf{m100} (movement+MORP on 100\% of masks).
However, for the expanded pools (\(N_{\text{synth}}{=}1804\) and \(3608\)), we exclusively employed the \textbf{m100} regime. Since the number of source seed patches is fixed ($\approx 902$), simply repeating the \textbf{nomove} generation would result in exact duplicates. Instead, we leveraged the stochastic nature of MORP (randomized apices, rays, and placement) to generate distinct geometric variations of the same source instances.
The \(N{=}1804\) and \(N{=}3608\) sets represent \(2\times\) and \(4\times\) stochastic expansions, respectively, where every synthetic mask possesses a unique geometry.
Full results appear in Table~\ref{tab:synth_morph_scale_multi_902_nomove} (\(N{=}902\)) and Table~\ref{tab:synth_morph_scale_multi_compare} (Expanded Pools).

The strongest configuration reached \textbf{60.07\%}~mIoU at \(N_{\text{synth}}{=}1804\), \(\lambda_{\text{synth}}{=}150\), and \textbf{m50}, outperforming the fine-tuned baseline on Peruvian data (54.07\% mIoU) by \textbf{+6.00~pp}. Class-wise improvements were concentrated in minority categories: \textit{oil} increased from 38.56\% to \textbf{49.37\%}~(+10.81~pp) and \textit{look-alike} from 23.37\% to \textbf{37.96\%}~(+14.59~pp), while dominant classes such as \textit{sea} and \textit{land} varied within~\(\pm1\)~pp.
\vspace{0.5em}
\begingroup
\setlength{\abovecaptionskip}{3pt}
\setlength{\belowcaptionskip}{0pt}

\begin{table*}[t]
\caption{Effect of synthetic \texttt{morph} (mask growth) and \texttt{scale} (synthetic dataset size) with multi-scale augmentation during mixed training (\textbf{ResNet-34 DeepLabV3+}, synth $N=902$). Values are IoU (\%).}
\centering
\resizebox{\textwidth}{!}{%
\begin{tabular}{c
cccc
cccc
cccc
cccc
cccc
cccc
}
\toprule
\textbf{Scale} & \multicolumn{4}{c}{mIoU} & \multicolumn{4}{c}{Sea} & \multicolumn{4}{c}{Oil} & \multicolumn{4}{c}{Look-alike} & \multicolumn{4}{c}{Ship} & \multicolumn{4}{c}{Land} \\
\cmidrule(lr){2-5}\cmidrule(lr){6-9}\cmidrule(lr){10-13}\cmidrule(lr){14-17}\cmidrule(lr){18-21}\cmidrule(lr){22-25}
 & nomove & m00 & m50 & m100 & nomove & m00 & m50 & m100 & nomove & m00 & m50 & m100 & nomove & m00 & m50 & m100 & nomove & m00 & m50 & m100 & nomove & m00 & m50 & m100 \\
\midrule
25  & 58.20 & 55.74 & 54.12 & 55.62 & 95.34 & 94.59 & 92.74 & 93.15 & 45.42 & 41.36 & 47.79 & 45.45 & 35.49 & 30.04 & 27.03 & 29.83 & 23.50 & 21.83 & 12.06 & 19.42 & 91.25 & 90.86 & 91.00 & 90.23 \\
50  & 57.83 & 56.14 & 56.01 & 54.69 & 94.90 & 94.52 & 94.61 & 93.34 & 50.31 & 44.23 & 44.03 & 47.17 & 33.99 & 29.26 & 29.06 & 27.63 & 18.31 & 22.81 & 22.54 & 16.37 & 91.62 & 89.89 & 89.81 & 88.93 \\
100 & 57.52 & 56.00 & 54.99 & 56.88 & 93.99 & 92.12 & 94.05 & 94.97 & 48.15 & 52.23 & 44.91 & 43.83 & 32.98 & 31.23 & 26.73 & 35.91 & 21.45 & 12.67 & 18.46 & 19.41 & 91.04 & 89.75 & 90.82 & 90.26 \\
150 & 56.89 & 54.68 & 54.55 & 57.43 & 93.58 & 94.80 & 94.86 & 95.04 & 52.12 & 40.12 & 38.01 & 47.07 & 30.69 & 27.55 & 26.21 & 32.16 & 19.22 & 21.24 & 24.27 & 23.08 & 88.83 & 89.70 & 89.40 & 89.82 \\
200 & 54.28 & 55.06 & 58.15 & 54.79 & 92.56 & 92.89 & 95.17 & 95.25 & 40.95 & 47.12 & 47.87 & 40.17 & 23.85 & 28.84 & 33.45 & 26.51 & 21.97 & 16.59 & 24.33 & 20.95 & 92.06 & 89.85 & 89.93 & 91.08 \\
\bottomrule
\end{tabular}}
\label{tab:synth_morph_scale_multi_902_nomove}
\end{table*}
\endgroup
\vspace{-4pt}  

\begingroup
\setlength{\abovecaptionskip}{3pt}
\setlength{\belowcaptionskip}{0pt}

\begin{table*}[t]
\caption{Comparison of IoU (\%) across classes for different synthetic scales using \texttt{m100}, synth $N = 1804$ and $N = 3608$ Peruvian training patches (ResNet-34 DeepLabV3+, mixed multi-scale training).}
\centering
\small
\resizebox{\textwidth}{!}{%
\begin{tabular}{c
cc
cc
cc
cc
cc
cc
}
\toprule
\textbf{Scale} 
& \multicolumn{2}{c}{\textbf{mIoU}} 
& \multicolumn{2}{c}{\textbf{Sea}} 
& \multicolumn{2}{c}{\textbf{Oil}} 
& \multicolumn{2}{c}{\textbf{Look-alike}} 
& \multicolumn{2}{c}{\textbf{Ship}} 
& \multicolumn{2}{c}{\textbf{Land}} \\
\cmidrule(lr){2-3}\cmidrule(lr){4-5}\cmidrule(lr){6-7}\cmidrule(lr){8-9}\cmidrule(lr){10-11}\cmidrule(lr){12-13}
 & 1804 & 3608 & 1804 & 3608 & 1804 & 3608 & 1804 & 3608 & 1804 & 3608 & 1804 & 3608 \\
\midrule
25  & 54.84 & 56.73 & 93.97 & 94.62 & 42.88 & 50.49 & 28.14 & 39.87 & 19.83 & 9.61  & 89.35 & 89.09 \\
50  & 53.04 & 57.25 & 94.48 & 96.00 & 37.80 & 39.83 & 23.91 & 32.22 & 17.61 & 24.97 & 91.39 & 93.23 \\
100 & 54.98 & 57.40 & 95.13 & 95.03 & 41.37 & 44.96 & 27.92 & 35.51 & 20.07 & 20.67 & 90.42 & 90.83 \\
150 & 60.07 & 56.73 & 94.83 & 93.41 & 49.37 & 51.19 & 37.96 & 31.72 & 27.30 & 16.10 & 90.90 & 91.21 \\
200 & 59.13 & 58.18 & 95.24 & 95.35 & 46.78 & 43.55 & 36.91 & 36.27 & 25.22 & 24.61 & 91.50 & 91.10 \\
\bottomrule
\end{tabular}}
\label{tab:synth_morph_scale_multi_compare}
\end{table*}
\endgroup
For the smallest synthetic pool (\(N_{\text{synth}}{=}902\)), performance generally rose with higher \(\lambda_{\text{synth}}\), peaking near 58.15\%~mIoU for \textbf{m50} at \(\lambda_{\text{synth}}{=}200\). Interestingly, the \textbf{nomove} control remained competitive even at low weights (58.20\% at \(\lambda_{\text{synth}}{=}25\)), indicating that additional synthetic supervision improves stability across the scale range. Expanding to \(N_{\text{synth}}{=}1804\) further stabilized minority-class IoUs and yielded the best overall means across mid–high weighting factors. Doubling again to \(N_{\text{synth}}{=}3608\) did not uniformly enhance accuracy: while strong \textit{oil} scores were preserved, \textit{ship} IoU tended to decrease, suggesting limited benefit from excessively large synthetic exposure when the underlying semantic diversity (number of unique seeds) remains constant.

Across all scales, moderate edits (\textbf{m50}) with mid high \(\lambda_{\text{synth}}\) provided the most consistent gains, while heavier modifications (\textbf{m100}) performed variably depending on pool size. Lighter regimes (\textbf{nomove}/\textbf{m00}) often remained competitive at smaller \(\lambda_{\text{synth}}\), confirming their effectiveness as stable baselines. These quantitative patterns are examined in detail in the Discussion.

Finally, running the same experiments on a higher-capacity \textit{Swin-Tiny UNet} yielded smaller or neutral changes (improvements in 3/12 settings; full results in Appendix \ref{tab:swin_unet_tiny_synth_morph_scale_multi}). This highlights architecture-dependent sensitivity to synthetic augmentation See Discussion ~\ref{sec:discussion}.

\subsection{Qualitative Results}

\paragraph{Visual Comparisons}

\begin{figure*}[t]
  \centering
  \resizebox{\textwidth}{!}{%
  \begin{tabular}{cccccc}
    \textbf{(a) SAR image} &
    \textbf{(b) GT mask} &
    \textbf{(c) (DLV3+)} &
    \textbf{(d) SwinUnetTiny} &
    \textbf{(e) FT-Synth 902} &
    \textbf{(f) FT-Synth 1804} \\
    \midrule

    \includegraphics[width=0.16\textwidth]{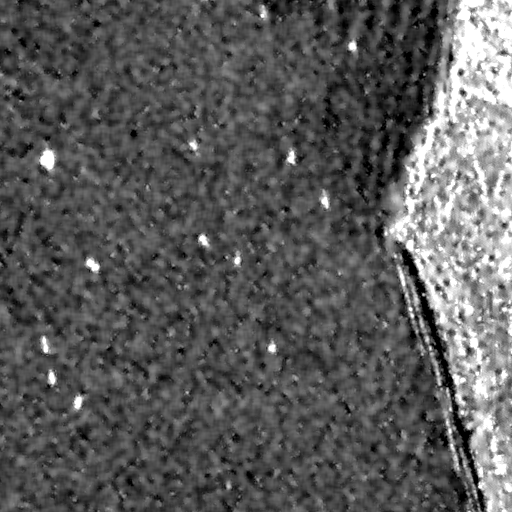} &
    \includegraphics[width=0.16\textwidth]{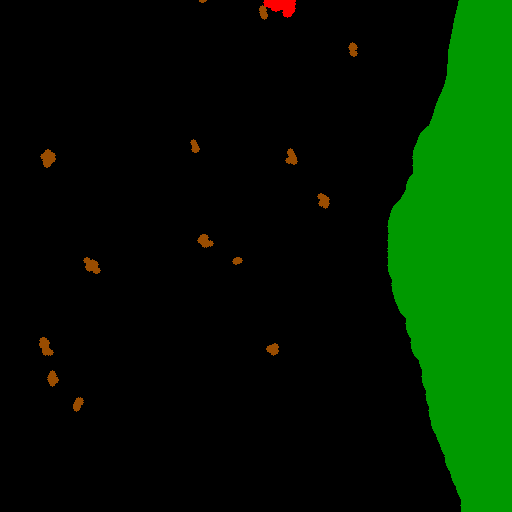} &
    \includegraphics[width=0.16\textwidth]{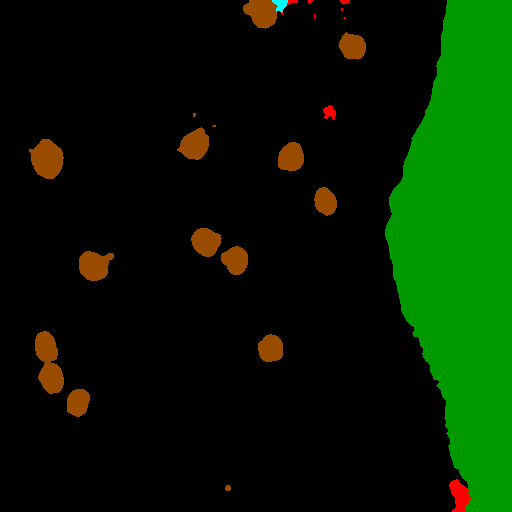} &
    \includegraphics[width=0.16\textwidth]{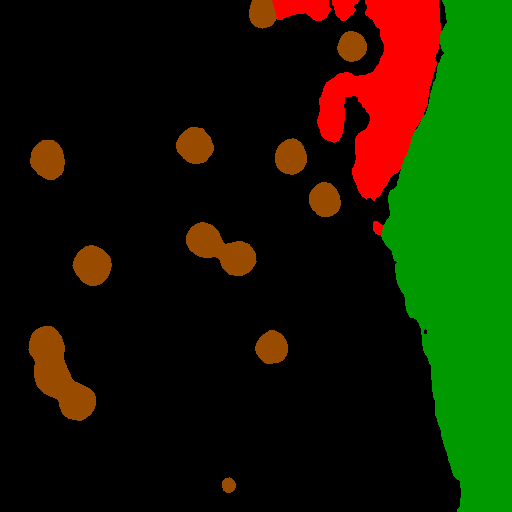} &
    \includegraphics[width=0.16\textwidth]{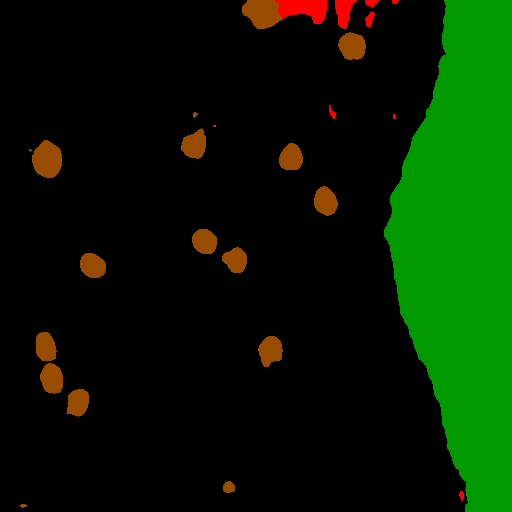} &
    \includegraphics[width=0.16\textwidth]{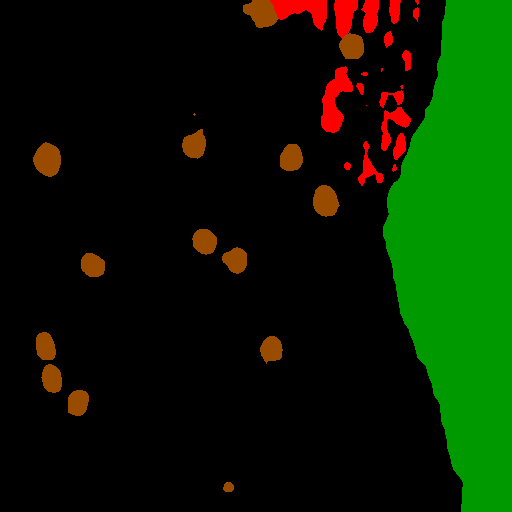} \\[0.6em]

    \includegraphics[width=0.16\textwidth]{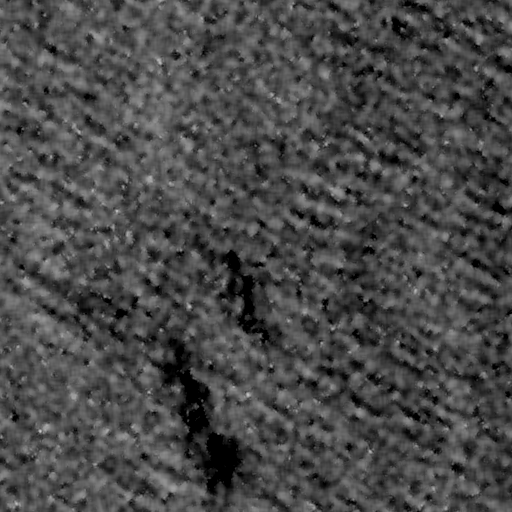} &
    \includegraphics[width=0.16\textwidth]{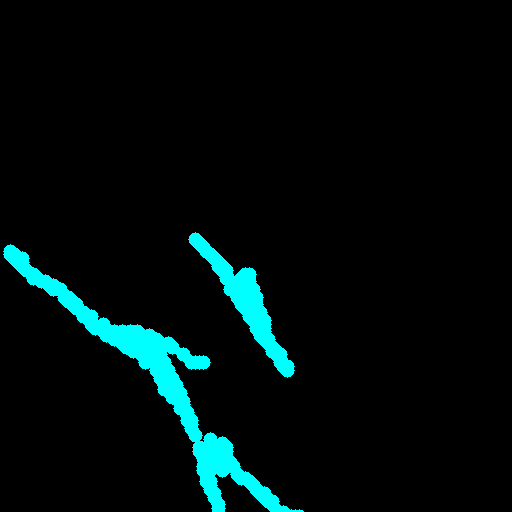} &
    \includegraphics[width=0.16\textwidth]{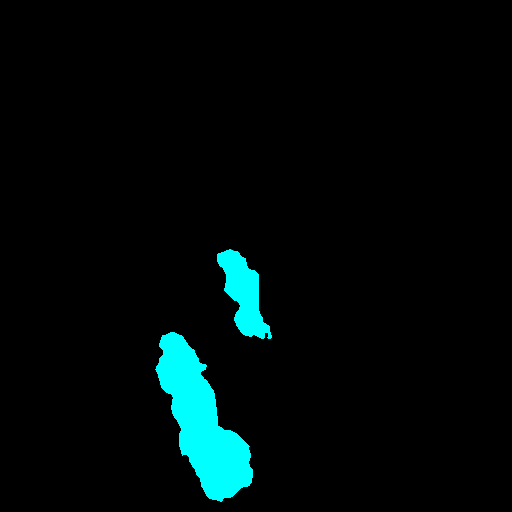} &
    \includegraphics[width=0.16\textwidth]{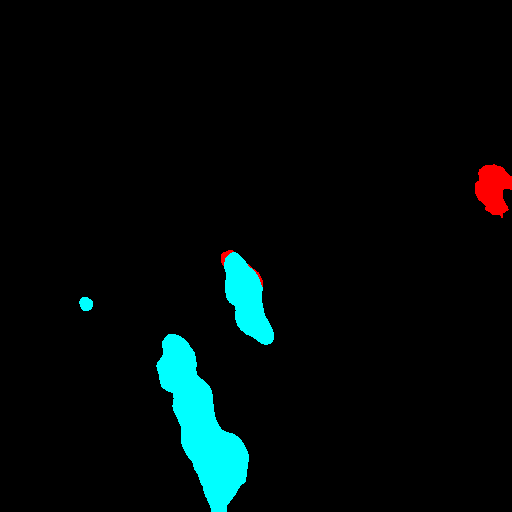} &
    \includegraphics[width=0.16\textwidth]{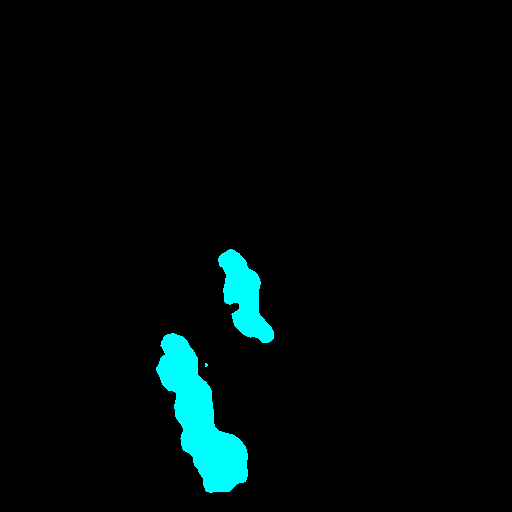} &
    \includegraphics[width=0.16\textwidth]{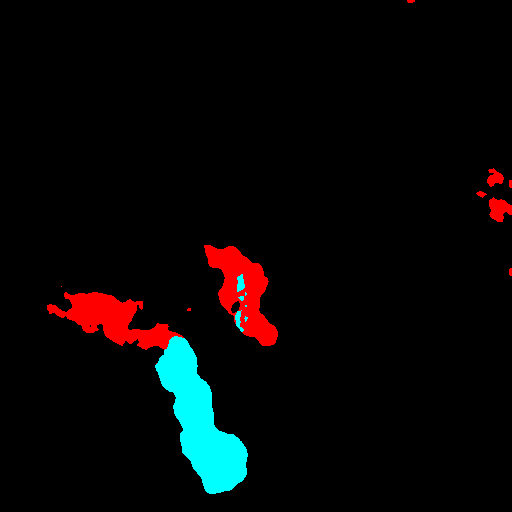} \\[0.6em]

    \includegraphics[width=0.16\textwidth]{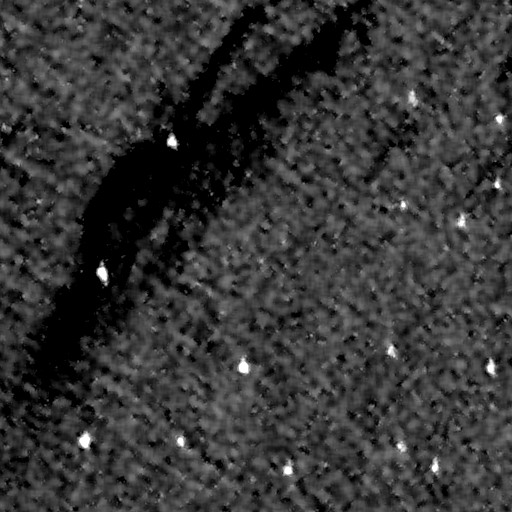} &
    \includegraphics[width=0.16\textwidth]{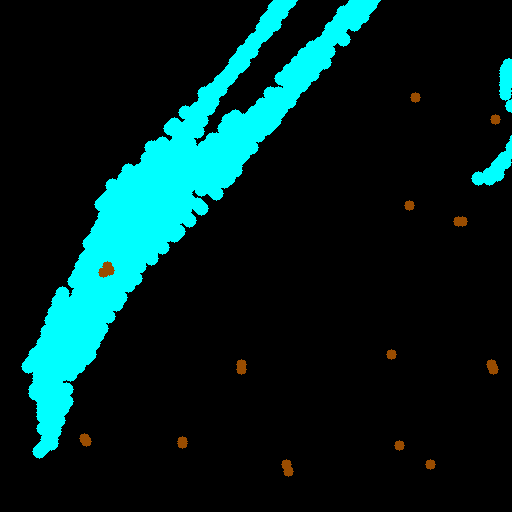} &
    \includegraphics[width=0.16\textwidth]{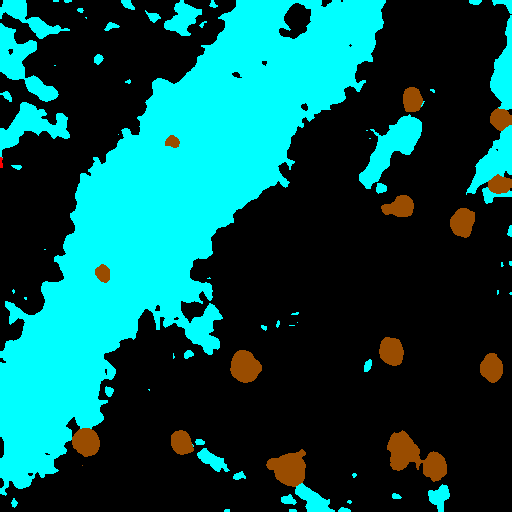} &
    \includegraphics[width=0.16\textwidth]{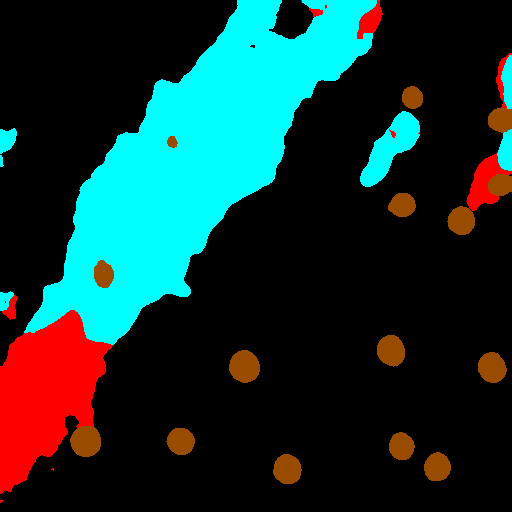} &
    \includegraphics[width=0.16\textwidth]{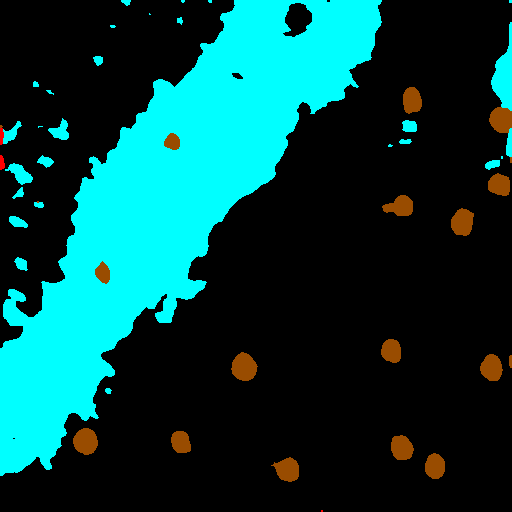} &
    \includegraphics[width=0.16\textwidth]{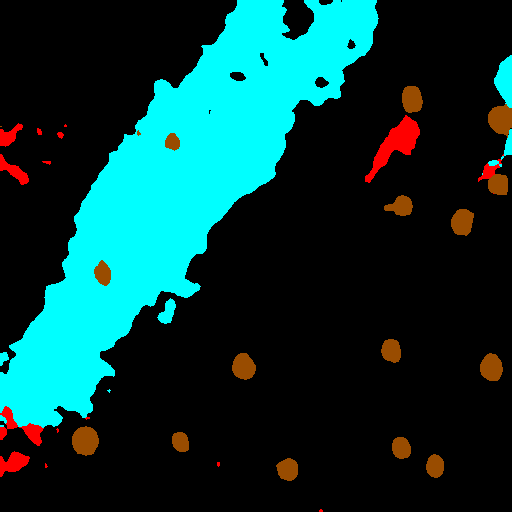} \\[0.6em]

    \includegraphics[width=0.16\textwidth]{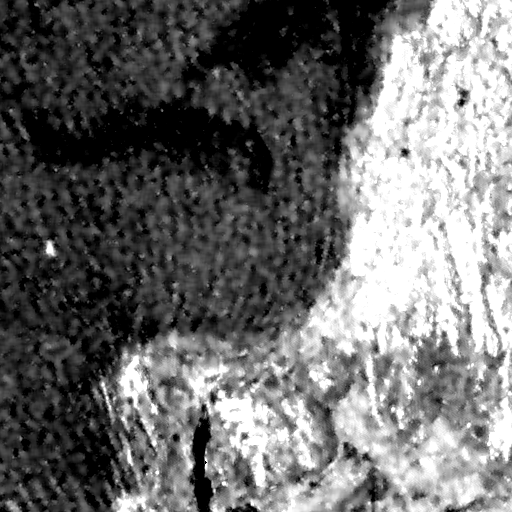} &
    \includegraphics[width=0.16\textwidth]{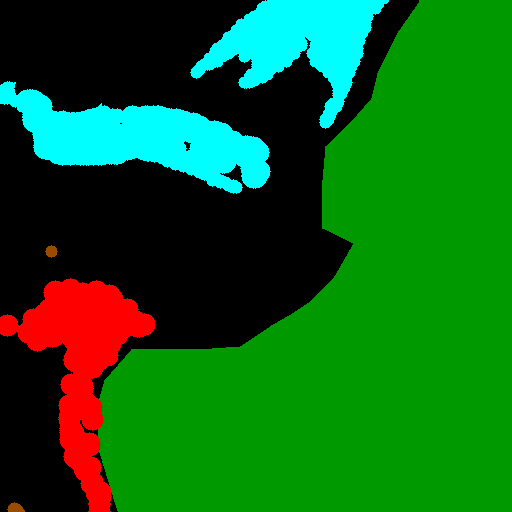} &
    \includegraphics[width=0.16\textwidth]{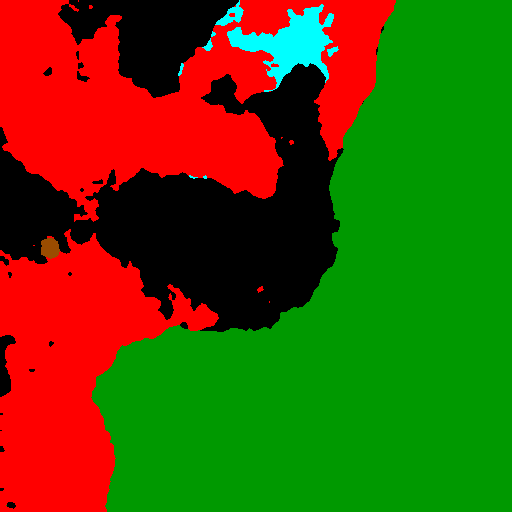} &
    \includegraphics[width=0.16\textwidth]{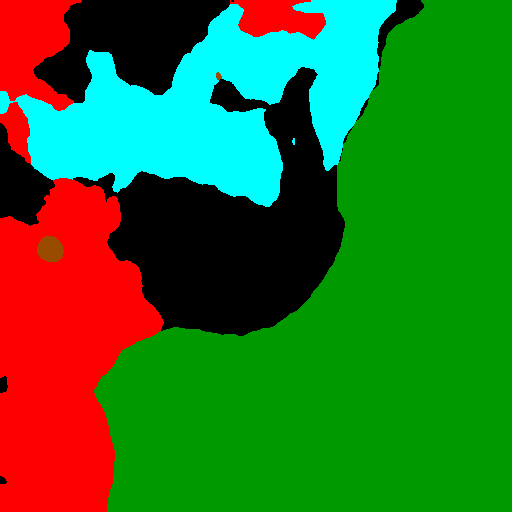} &
    \includegraphics[width=0.16\textwidth]{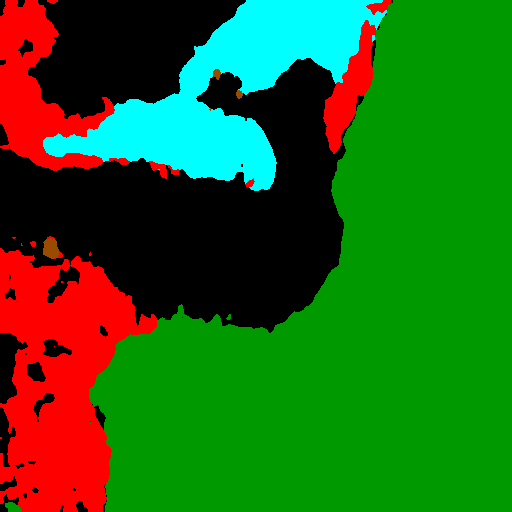} &
    \includegraphics[width=0.16\textwidth]{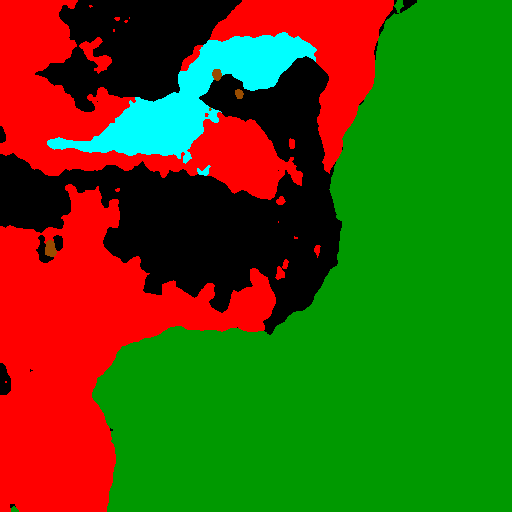} \\[0.6em]

    \includegraphics[width=0.16\textwidth]{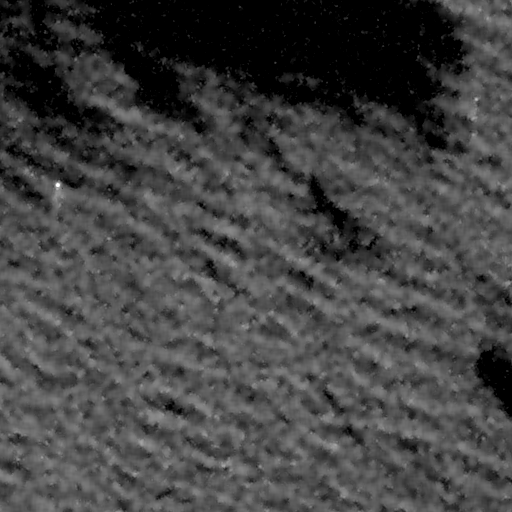} &
    \includegraphics[width=0.16\textwidth]{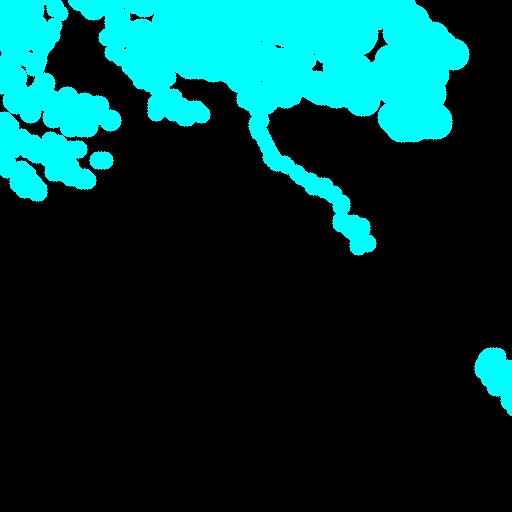} &
    \includegraphics[width=0.16\textwidth]{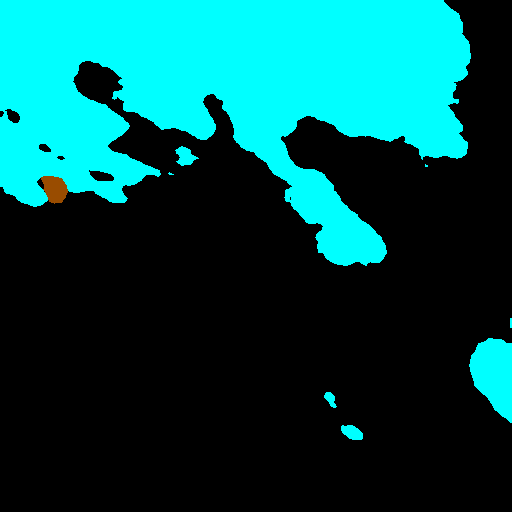} &
    \includegraphics[width=0.16\textwidth]{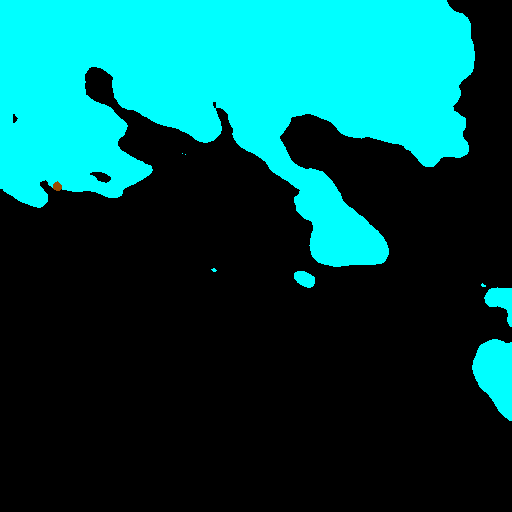} &
    \includegraphics[width=0.16\textwidth]{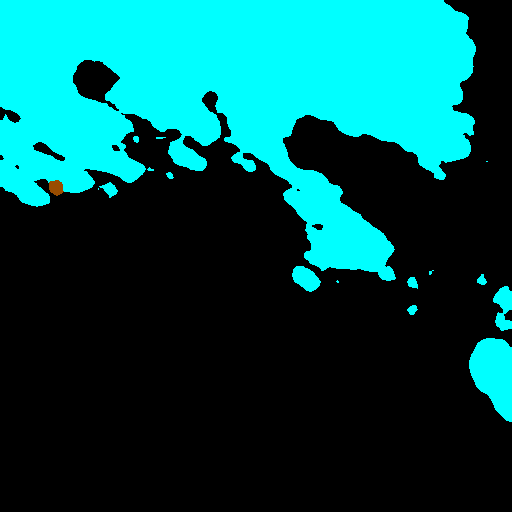} &
    \includegraphics[width=0.16\textwidth]{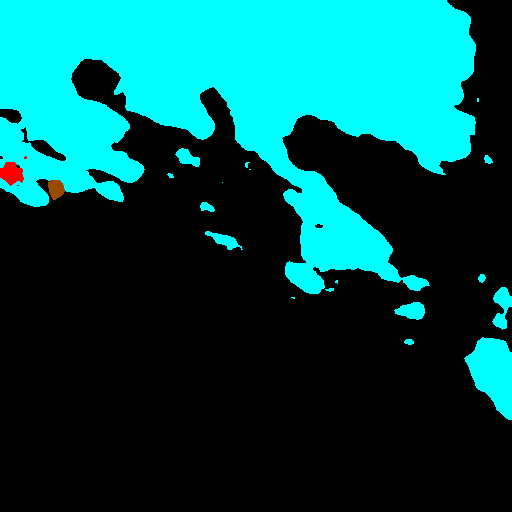} \\

  \end{tabular}}
  \caption{Comparison of oil spill segmentation predictions across five Peruvian Sentinel-1 patches. Columns show: (a) SAR input, (b) ground truth mask, (c–d) fine-tuned baselines (DeepLabV3+, SwinUnetTiny), and (e–f) fine-tuned models with synthetic data using $N{=}902$ and $N{=}1804$ real samples. Colors follow the palette defined in Sec.~\ref{sec:study-area-datasets}).}
  \label{fig:patch_comparison_updated}
\end{figure*}

Figure~\ref{fig:patch_comparison_updated} presents qualitative outputs on representative Peruvian Sentinel-1 patches for (c) \textit{DeepLabV3\texttt{+}}, (d) \textit{Swin-Tiny UNet}, and their synthetically fine-tuned counterparts with $N{=}902$ and $N{=}1804$ real samples (e–f). At patch scale, both baselines delineate the main slick bodies but occasionally fragment diffuse perimeters and over-respond to textured sea. Synthetic fine-tuning smooths contours and suppresses small spurious islands, with the $N{=}902$ variant giving the most complete outlines on thin, wispy edges, and the $N{=}1804$ variant favoring more cautious responses near complex background. We therefore use $N{=}1804$ for full-scene panels below to emphasize low oil false positives, and $N{=}902$ in patch panels to illustrate boundary completeness.

\paragraph{Per-Scene Area Statistics}
Table~\ref{tab:scene_areas_by_img} summarizes per-scene IoU and predicted/false-positive areas for \textit{ResNet-34 DeepLabV3\texttt{+}} with \textbf{FT–Synth~1804} over eight test scenes ($5567\times5567$ at 10\,m). Slick footprints span two orders of magnitude (e.g., scene~\texttt{14} $\approx 47.7$\,km$^2$ vs.\ scene~\texttt{18} $\approx 0.43$\,km$^2$), and look-alike coverage can dominate (e.g., \texttt{17}, \texttt{19}, \texttt{21}), reflecting wind-slick bands, biogenic films, internal-wave striations, and rain-cell signatures typical of the Peruvian coast. Under this variability, \textbf{FT–Synth~1804} maintains low oil false positives in half the scenes (FP-oil $\leq 0.31$\,km$^2$ in \texttt{17}, \texttt{18}, \texttt{21}, \texttt{25}; zero in \texttt{18}), while higher FP-oil concentrates in cluttered nearshore or frontal zones (\texttt{14}, \texttt{19}, \texttt{26}, \texttt{28}). IoU co-varies with footprint size and background complexity (e.g., \texttt{28}: 0.713; \texttt{19}: 0.620; \texttt{14}: 0.579), consistent with larger, well-contrasted slicks being easier to segment than thin filaments embedded in structured sea state. These per-scene numbers ground the qualitative panels by quantifying the operationally relevant trade-off: strong boundary adherence where contrast supports it, and restrained oil alarms where background phenomena dominate.

\begin{figure*}[t]
  \centering
  \begin{tabular}{cc}
    \includegraphics[width=0.45\textwidth]{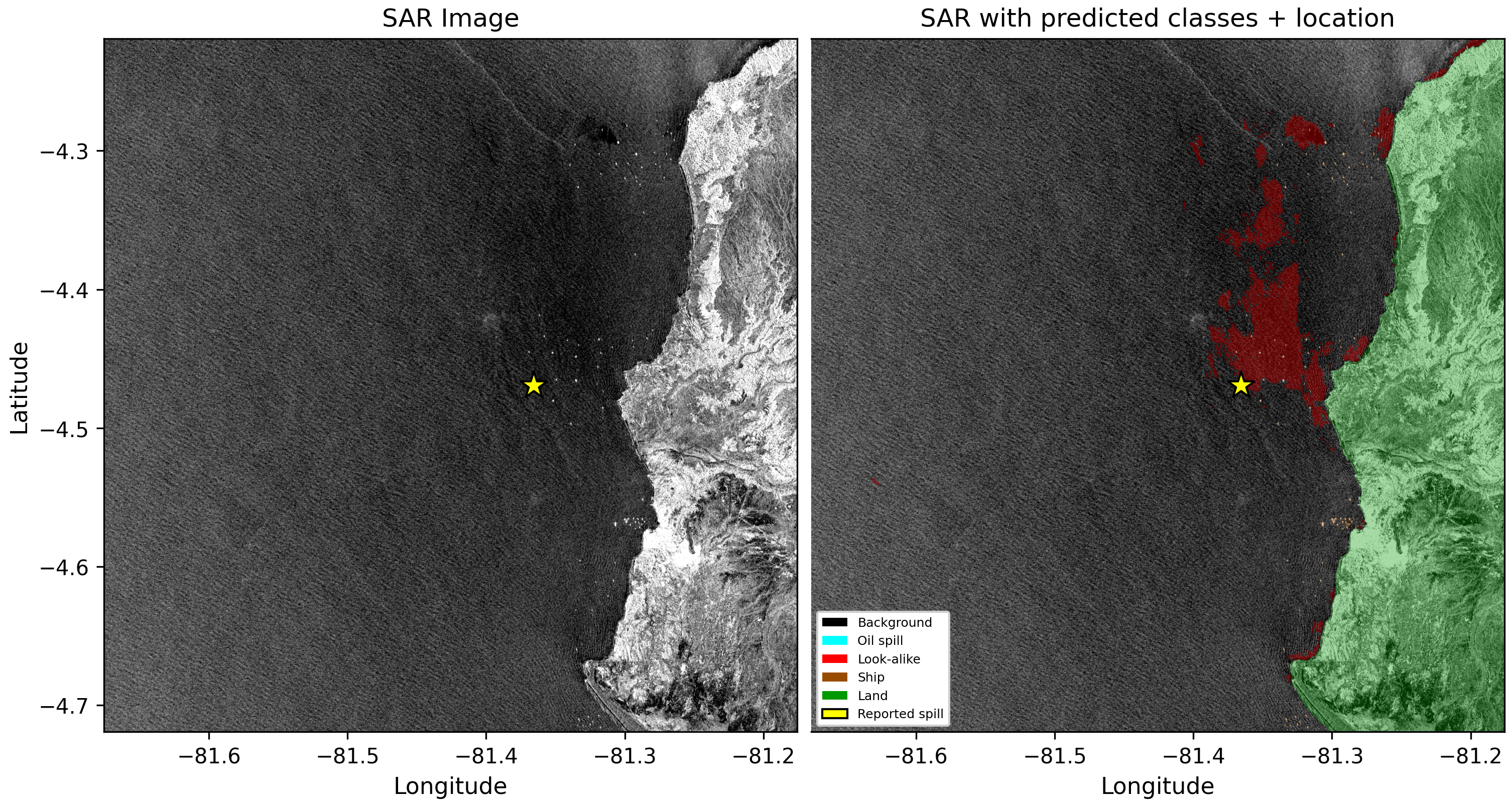} &
    \includegraphics[width=0.45\textwidth]{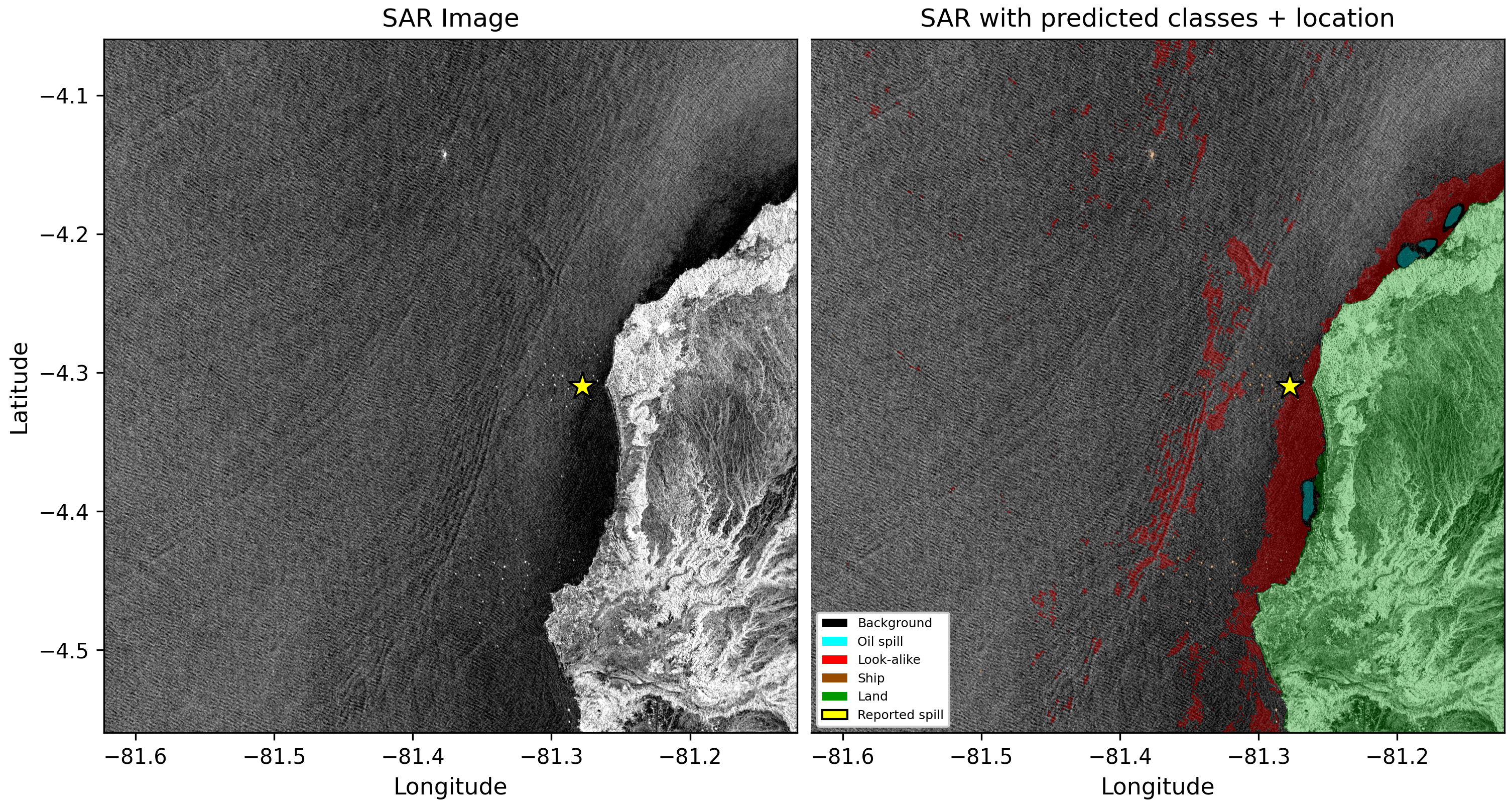} \\
    (a) Img 18 & (b) Img 19 \\[1em]

    \includegraphics[width=0.45\textwidth]{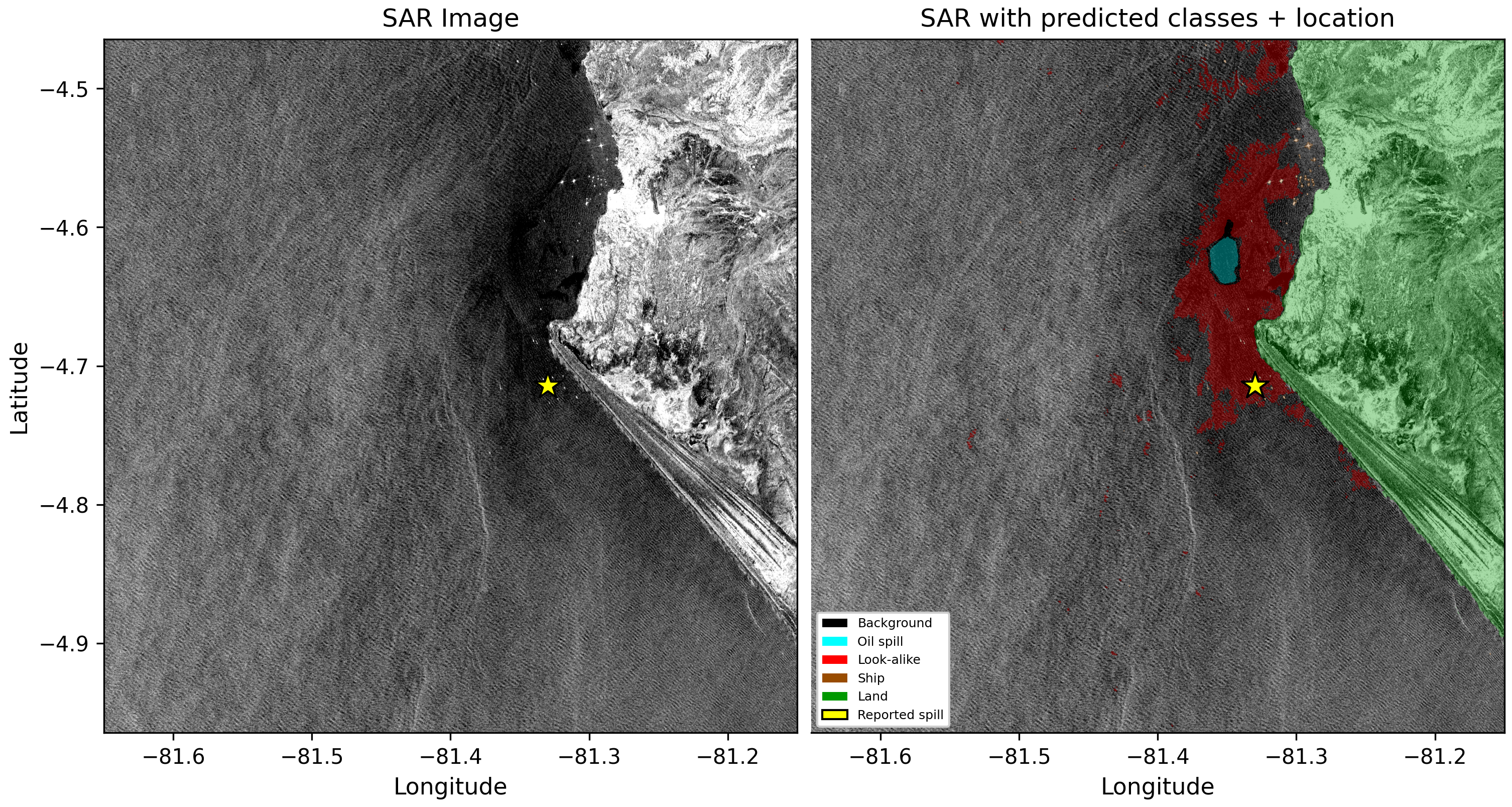} &
    \includegraphics[width=0.45\textwidth]{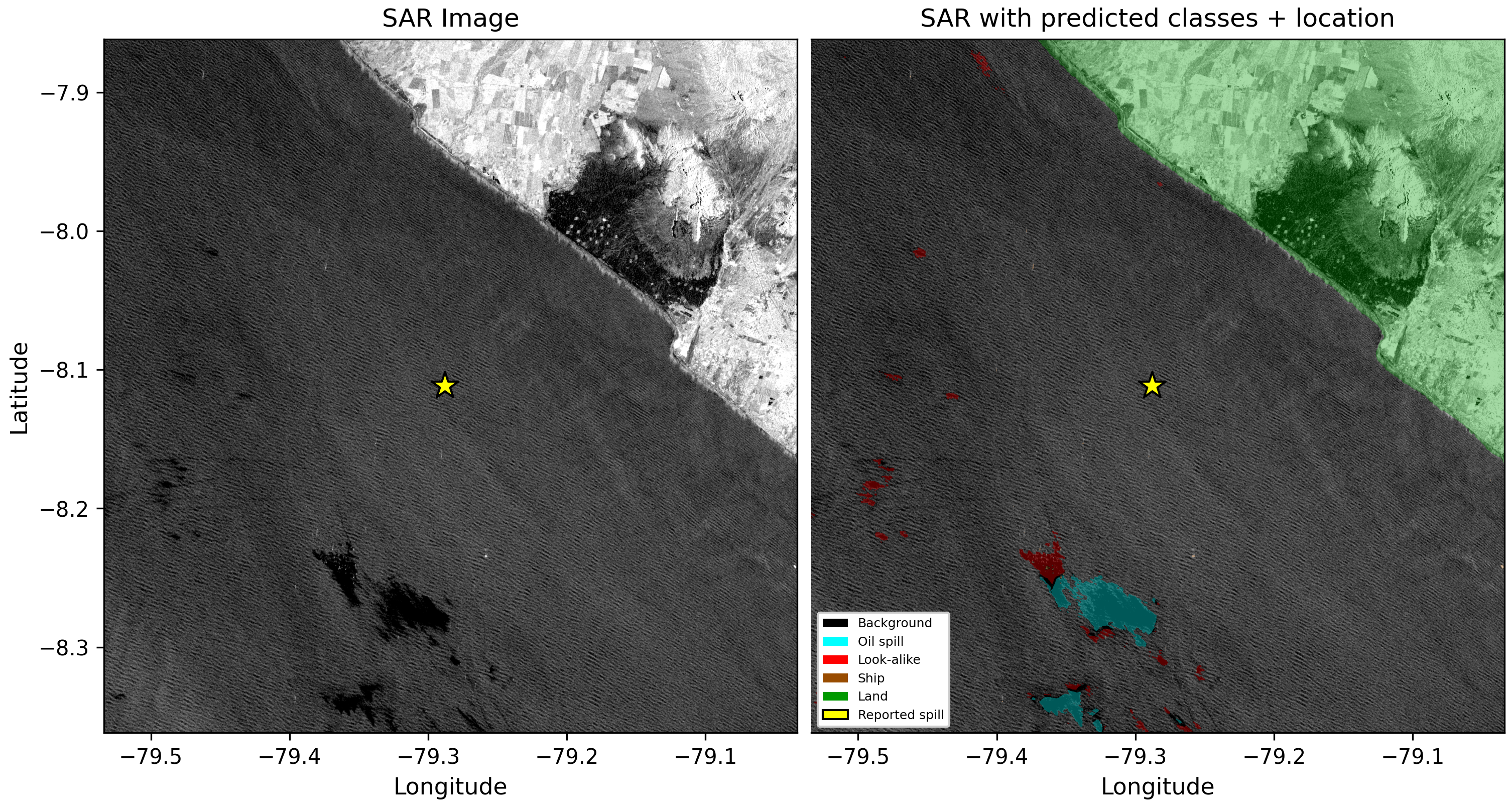} \\
    (c) Img 26 & (d) Img 14 \\
  \end{tabular}

  \caption{Full‐image segmentation examples on four Peruvian spill test scenes. These scenes cover a range of sea states, coastline geometries, and contrast conditions. Corresponding per-scene IoU and predicted/false-positive areas are reported in Appendix Table~\ref{tab:peruvian_selection}.}
  \label{fig:full_seg_examples}
\end{figure*}

\paragraph{Full-Scene Qualitative Panels}
Figure~\ref{fig:full_seg_examples} shows full image segmentations of four test scenes using \textbf{FT–Synth~1804}. 
(a) Img~18, small offshore sheen with weak contrast amid low-wind background; the model avoids spurious oil, consistent with FP-oil $\approx 0$\,km$^2$. 
(b) Img~19, broad look-alike bands (likely wind/internal-wave structure) with embedded slick fragments; predictions follow the main cyan plume while limiting coastal false alarms (IoU $\approx 0.62$). 
(c) Img~26, nearshore edge with bright backscatter from breaking waves and possible layover/foreshortening; the model retains the along-shelf slick but is conservative against shoreline-adjacent patches where rain cells and surf texture increase confusion. 
(d) Img~14, large, high-contrast event with complex coastline; segmentation recovers the dominant extent and preserves land boundaries, with residual false positives appearing in frontal zones where sea-state transitions are sharp (IoU $\approx 0.58$). 
Across these scenes the outputs remain spatially coherent: continuous slick axes align with ambient texture gradients, vessel wakes are not mislabelled as slicks, and dark streaks from environmental modulation are largely filtered. Together with Table~\ref{tab:scene_areas_by_img}, these panels illustrate that the model scales from patch-level detail to full-scene mapping under heterogeneous sea states while keeping oil false positives operationally low.

\begin{table}[htbp]
\caption{Per-scene IoU and predicted/false-positive areas (km\(^2\)) using FT–Synth~1804, indexed by image ID from Table~\ref{tab:peruvian_selection} in Appendix.}
\centering
\resizebox{\columnwidth}{!}{%
\begin{tabular}{lrrrrr}
\toprule
\textbf{Scene ID} & \textbf{mIoU} & \textbf{Pred.\ oil (km\(^2\))} & \textbf{Pred.\ look-alike (km\(^2\))} & \textbf{FP oil (km\(^2\))} & \textbf{FP look-alike (km\(^2\))} \\
\midrule
14 & 0.5793 & 29.5586 & 60.1318 & 10.2360 & 60.1318 \\
17   & 0.4521 & 0.0018  & 232.5382 & 0.0018  & 199.9045 \\
18   & 0.4522 & 0.0000  & 88.1186  & 0.0000  & 86.3783  \\
19  & 0.6201 & 6.2375  & 182.0464 & 6.2375  & 109.3411 \\
21  & 0.5334 & 0.3965  & 195.4687 & 0.1922  & 195.4687 \\
25  & 0.5333 & 0.3096  & 154.2089 & 0.3096  & 153.9361 \\
26  & 0.4427 & 6.5551  & 142.2415 & 4.6652  & 137.7332 \\
28  & 0.7126 & 3.4302  & 37.1061  & 1.0267  & 37.1061  \\
\bottomrule
\end{tabular}
}
\label{tab:scene_areas_by_img}
\end{table}

\subsubsection{Temporal evolution of major spill events in peruvian coast.}
\begin{figure*}[t]
  \centering
  \begin{tabular}{cc}
    \includegraphics[width=0.45\textwidth]{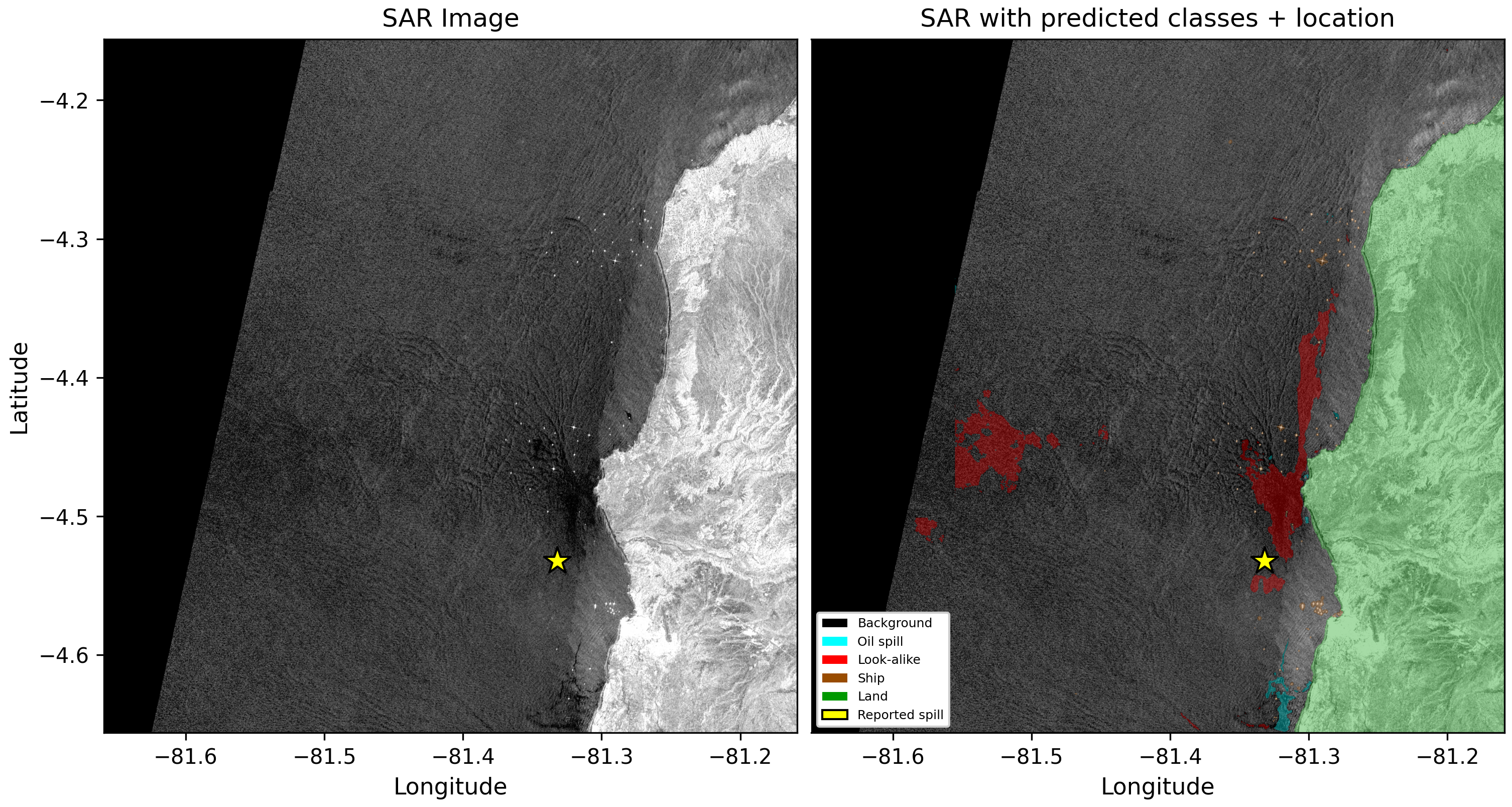} &
    \includegraphics[width=0.45\textwidth]{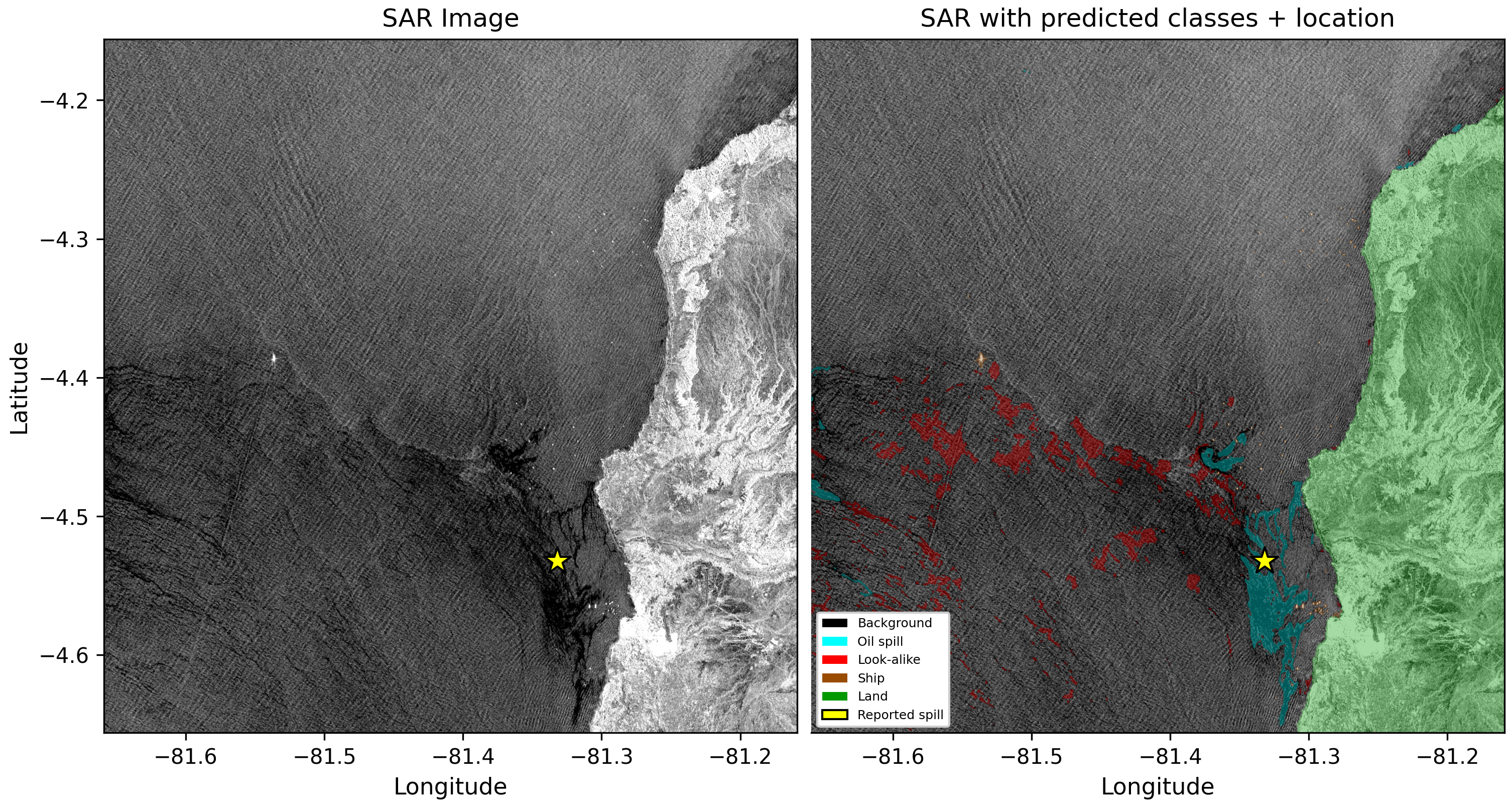} \\
    
    (a) & (b) \\[1em] 

    \includegraphics[width=0.45\textwidth]{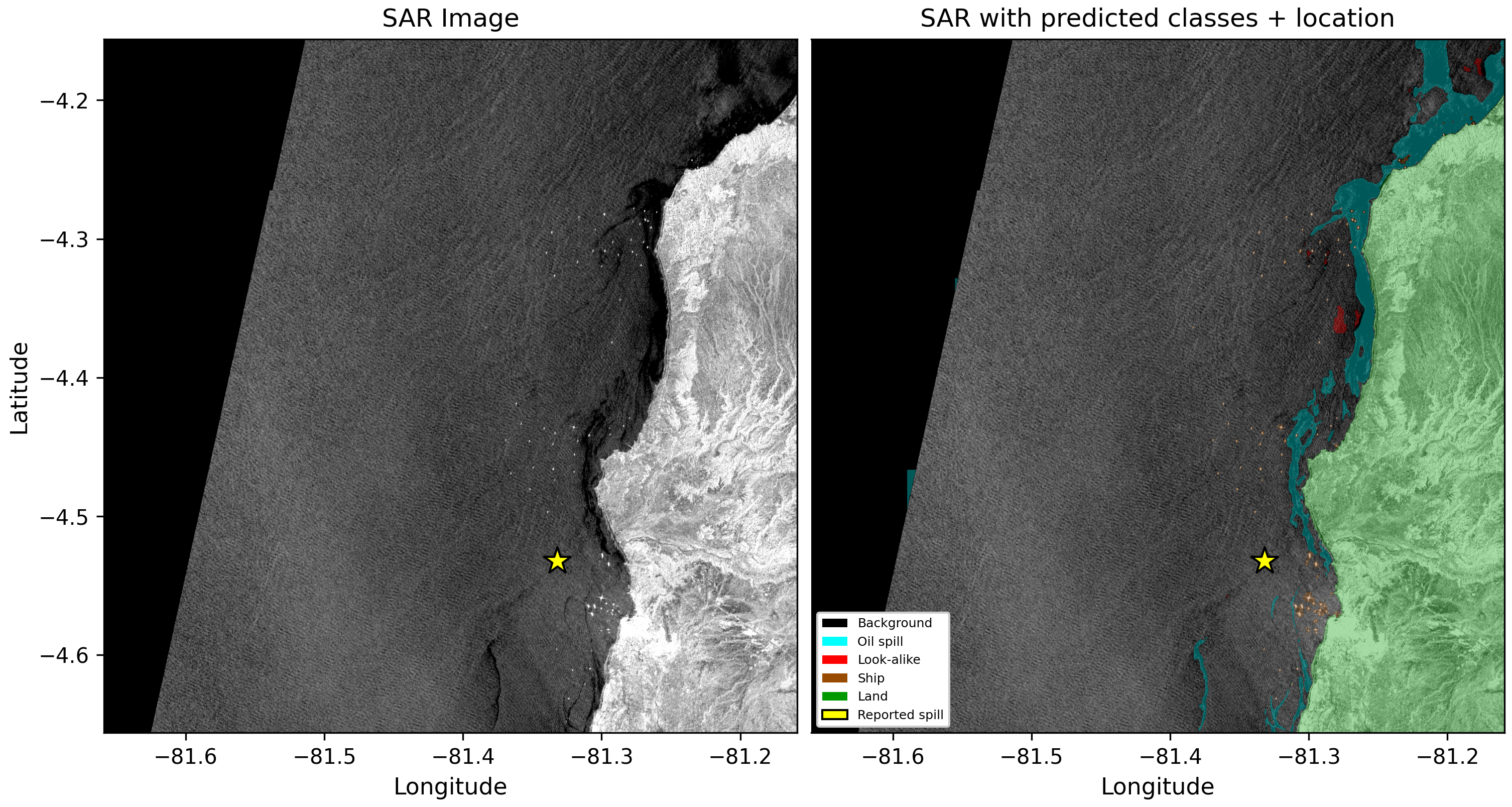} &
    \includegraphics[width=0.45\textwidth]{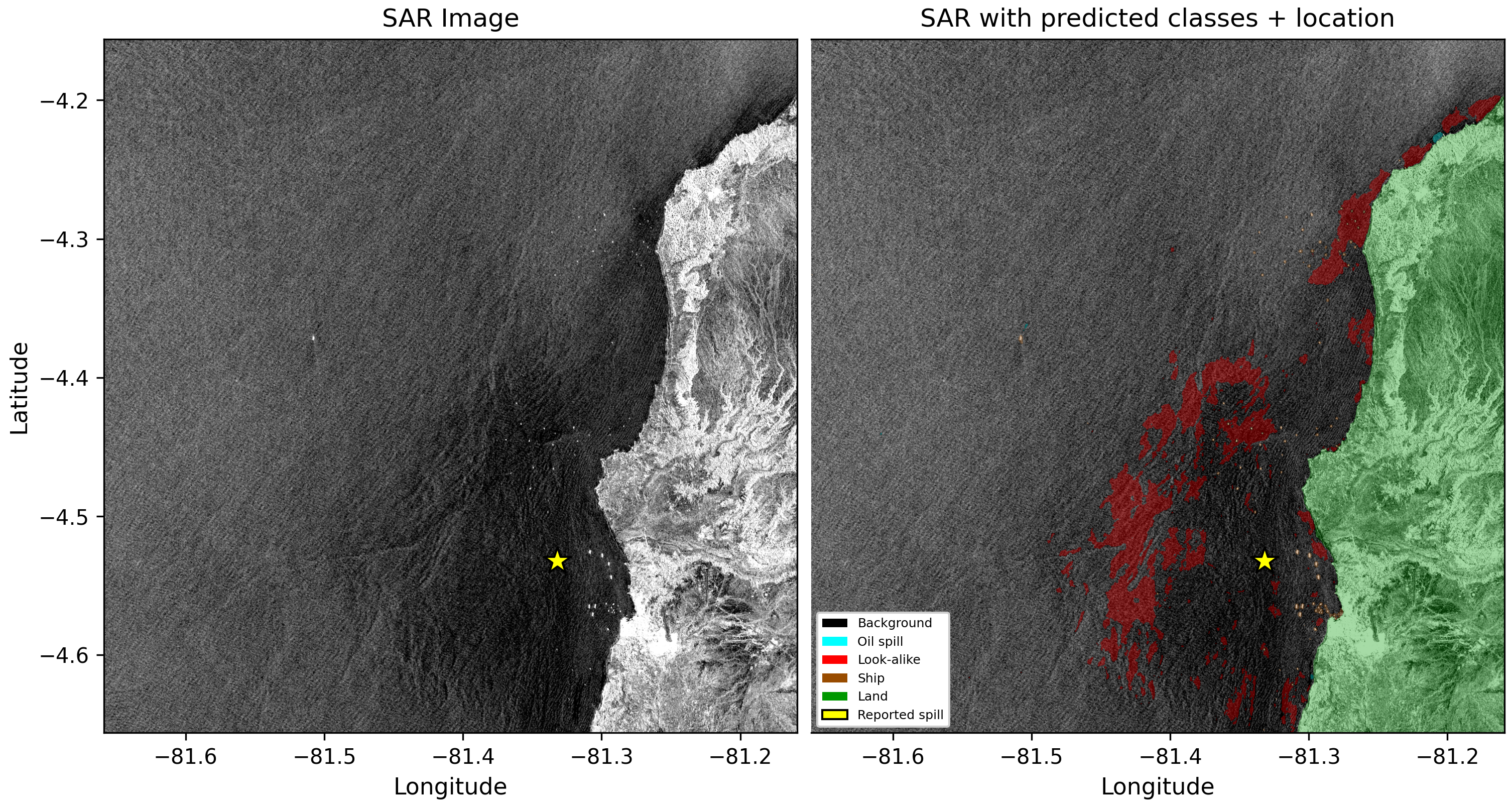} \\
    
    (c) & (d) \\
  \end{tabular}

  \caption{Lobitos offshore spill sequence (26 Dec 2024–10 Jan 2025) as segmented by the FT–Synth~1804 model. 
  For each date, the left panel shows Sentinel-1 VV backscatter and the right panel shows predicted classes. The yellow star marks the reported discharge location. Panels correspond to image IDs (a) 38, (b) 39, (c) 37, and (d) 40 (see Appendix Table~\ref{tab:peruvian_selection} for reference).}
  \label{fig:lobitos_temporal}
\end{figure*}

\begin{figure}[t]
    \centering
    \begin{tabular}{cccccc}
    \includegraphics[width=\columnwidth]{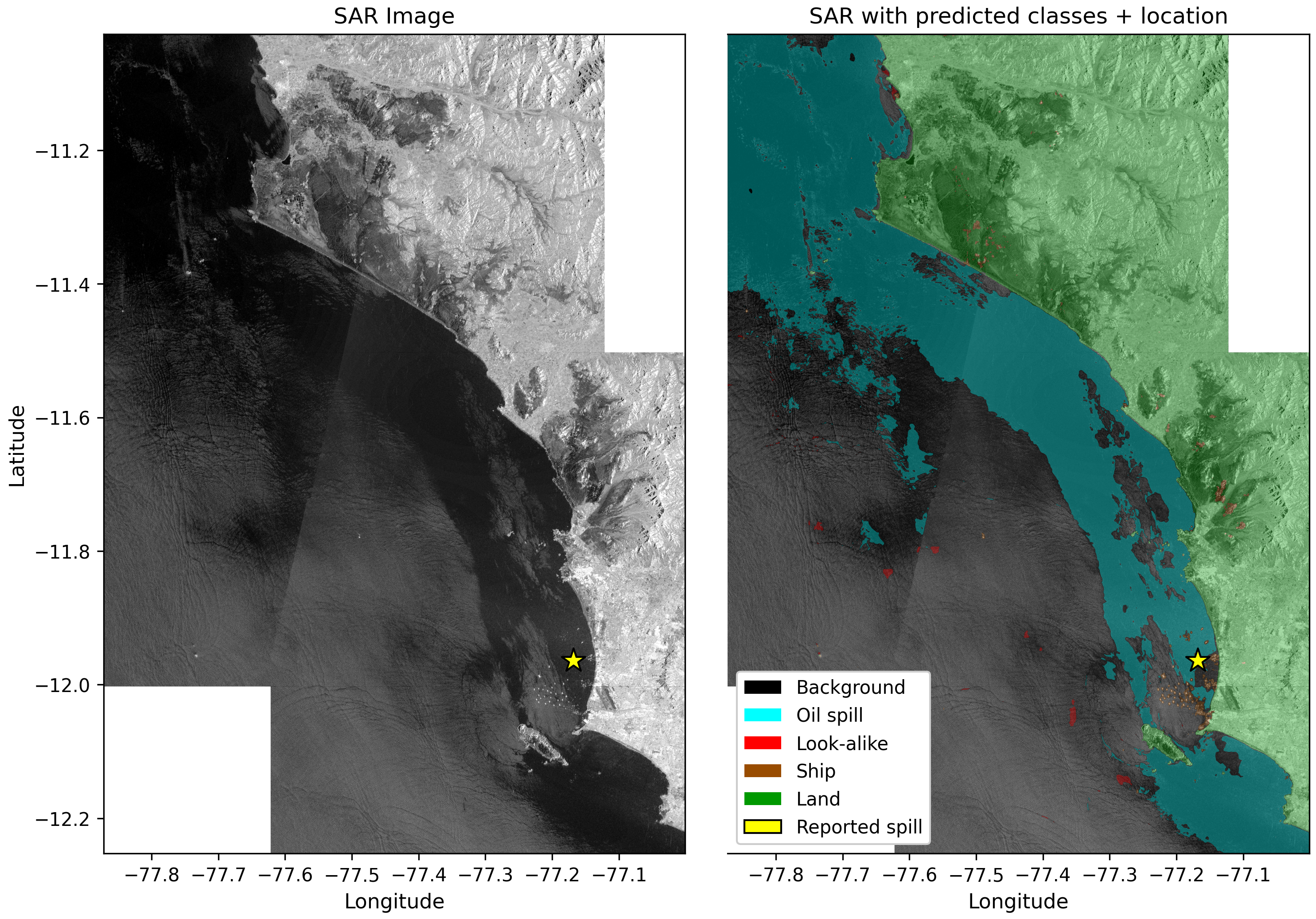}
    \end{tabular}
    \caption{Sentinel-1 scene from 25 Jan 2022, the first SAR acquisition covering the Repsol/Ventanilla spill. 
    Left: VV backscatter mosaic. Right: FT–Synth~1804 predictions overlaid on SAR image. The yellow star marks the reported discharge location.
    Image ID 30 (see Appendix Table~\ref{tab:peruvian_selection}).}
    \label{fig:repsol_jan25}
\end{figure}

\begin{figure}[t]
    \centering
  \begin{tabular}{cccccc}
    \includegraphics[width=\columnwidth]{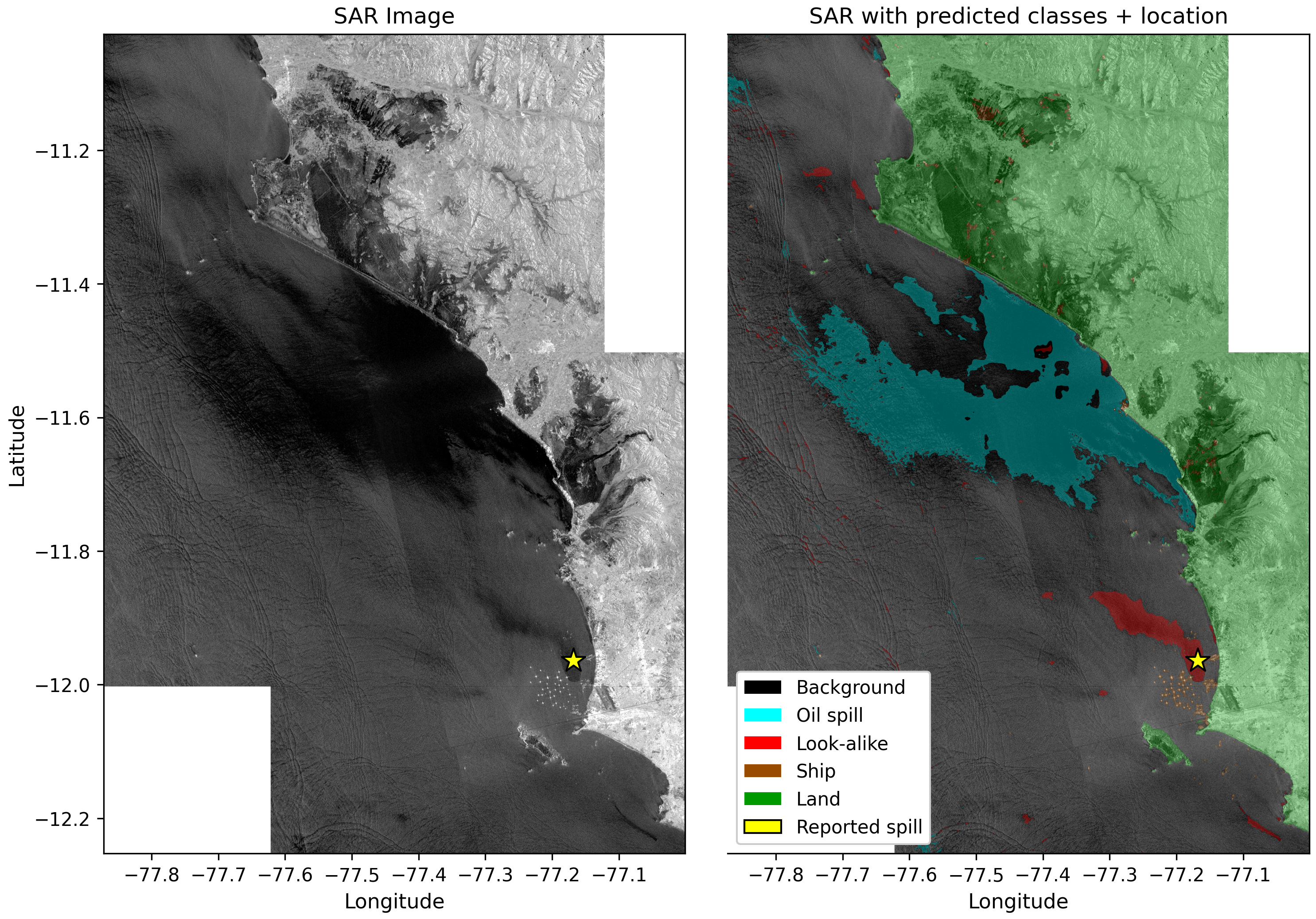}
    \end{tabular}
    \caption{Sentinel-1 scene from 02 Feb 2022, showing the northward dispersion of the Repsol/Ventanilla spill toward Ancon. 
    Left: VV backscatter mosaic. Right: FT–Synth~1804 predictions overlaid on SAR image. The yellow star indicates the nominal discharge location. 
    Image ID 29 (see Appendix Table~\ref{tab:peruvian_selection}).}
    \label{fig:repsol_feb02}
\end{figure}

Two high-impact events were examined to assess the temporal stability of the fine-tuned MORP-Synth model. 
Although included in training, the Lobitos offshore spill sequence (26 Dec 2024–10 Jan 2025; Fig.~\ref{fig:lobitos_temporal}) 
and the Repsol/Ventanilla event (25 Jan–02 Feb 2022; Fig.~\ref{fig:repsol_feb02},~\ref{fig:repsol_jan25}) serve as temporal plausibility checks. 

\paragraph{Lobitos sequence.}
The \textit{Lobitos sequence} (26 Dec 2024–10 Jan 2025; Fig.~\ref{fig:lobitos_temporal}). evaluates the model's ability to track the same spill through time. On 26~December, the segmentation captures a compact slick of about 5.2~km$^2$ near the source, although part of the adjacent elongated anomaly is labelled as look-alike rather than oil. Its dark, coherent appearance suggests a thin sheen, indicating a likely false negative arising from the model's conservative treatment of low-contrast filaments. 

By 29~December, the detected oil area increases to roughly 40~km$^2$, accompanied by about 76~km$^2$ of look-alike surface modulation. The 7~January scene shows northward transport and fragmentation, reaching an estimated 86~km$^2$ of oil with limited look-alike interference. By 10~January, only faint streaks remain (<1~km$^2$ of oil against 144~km$^2$ of look-alike), consistent with dispersion and thinning below SAR detectability. Overall, the sequence illustrates coherent temporal evolution, initial localization, spreading, and eventual decay—with the early false negative reflecting the inherent difficulty of segmenting marginal slicks under low contrast.

\paragraph{Repsol/Ventanilla sequence.}

The Repsol/Ventanilla event (25~January–02~February~2022; Figs.~\ref{fig:repsol_jan25} and~\ref{fig:repsol_feb02}) provides a second test of temporal behavior over a much larger spatial domain than the Lobitos case. Each mosaic spans roughly $1.3\times10^{4}~\mathrm{km}^{2}$, with complex radiometric structure from surf, shallow-water roughness and cloud shadowing. On 25~January (Fig.~\ref{fig:repsol_jan25}), the model detects an elongated slick of \(2.93\times10^{3}\,\mathrm{km}^{2}\) over the full mosaic, extending northward from the reported discharge location. South of the yellow reference point, however, large areas are labelled as oil despite official assessments indicating that the spill did not propagate into this sector; these southern detections should therefore be interpreted as false positives linked to coastal darkening and wave shadow effects. 

By 2~February (Fig.~\ref{fig:repsol_feb02}), the model identifies a coherent plume displaced northward toward Ancón, with the oil-labelled area decreasing to about $1\,143~\mathrm{km}^{2}$, consistent with surface transport and thinning documented in hydrodynamic reconstructions of the event ~\cite{mogollonREPSOLOilSpill2023}. In this second frame, a compact red patch near the coast appears to be a false negative (oil misclassified as look-alike), as its geometry and tonal contrast resemble the main slick rather than a natural surfactant feature. Despite these local misclassifications, the model preserves the alongshore continuity of the plume, avoids severe fragmentation, and follows a realistic northward drift between the two dates. As in the Lobitos sequence, the temporal evolution of the predicted slicks remains smooth and physically plausible.

\section{Discussion}
\label{sec:discussion}
\paragraph{Overview of Main Findings}

Our research demonstrates that geometrically diversified synthetic augmentation (MORP-Synth) substantially reduces the cross-domain performance gap between Mediterranean source data and the Peruvian target environment. While standard fine-tuning provides initial adaptation, it struggles with the severe class imbalance inherent to oil spill datasets. The proposed augmentation strategy specifically targeted this limitation. The strongest configuration ($N{=}1804$, $\lambda_{\text{synth}}{=}150$, moderate edits) delivered a 6.00 pp improvement in overall mIoU compared to the fine-tuned baseline. Most critically, this gain was driven by a substantial recovery of minority classes: Oil Spill IoU increased by +10.81 pp and Look-alike IoU by +14.59 pp. This indicates that synthetic diversity is particularly effective at resolving the confusion between low-contrast slicks and naturally occurring dark formations (wind shadows)~\cite{wanStudyVariationPatterns2024,garcia-pinedaClassificationOilSpill2020}. We also observed an architecture-dependent response: mid capacity CNNs (ResNet-DeepLabv3+) benefited significantly from synthetic regularization, whereas the Swin-Tiny UNet, which already had strong baseline performance, showed smaller or neutral gains under MORP–Synth. Beyond these immediate performance gains, the recursive nature of the mask-based perturbations establishes a scalable framework for continuous data generation, offering a systematic solution to the chronic data scarcity that typically bottlenecks CNN generalization.

By systematically discovering and perturbing boundary apices and shifting object placement (Stage A of MORP), we decoupled the slick's geometry from its radiometric texture. Since the GAN generator (INADE) maintains the spectral properties of SAR backscatter while the mask geometry changes, the network is forced to learn robust textural features (dampening ratios, gradient boundaries) rather than overfitting to specific spill contours. This explains why the \textit{nomove} baseline (which adds texture diversity but no shape diversity) was less effective than the \textit{m50} regime: geometry appearance decoupling appears particularly effective in our experiments to break the semantic confusion between oil and look-alikes ~\cite{wangSinGANLabelerEnhancedSinGAN2025}. Conversely, the performance drop at \textit{m100}consistent with the idea that overly aggressive curvature edits may produce shapes that are less physically plausible for wind and current driven slicks, which in practice acts as label noise.

\paragraph{Interpretation and Comparison with Previous Studies}
The observed cross-domain gap and its partial closure through fine-tuning are consistent with prior SAR segmentation studies that emphasize local adaptation before operational deployment~\cite{charngDeepLearningSegmentation2020,dehghani-dehcheshmehOilSpillsDetection2023,cuiEnhancedUnsupervisedDomain2025}. The faster convergence and better endpoints of fine-tuned models relative to ImageNet-initialized runs align with the view that marine backscatter statistics learned on one region offer transferable mid-level priors even under shifted incidence-angle distributions and sea-state regimes~\cite{pratapFineArtFinetuning2025,kussulTransferLearningSinglepolarized2025}. Regarding augmentation, our results suggest that the proposed pipeline improves generalization by decoupling geometric priors from textural signatures. In standard training, CNNs often overfit to the specific elongation or smooth boundaries typical of oil slicks (geometric priors). MORP (Stage A) disrupts this correlation by introducing high-variance, stochastically perturbed shapes via curvature driven mask editing, since the conditional generator (Stage B) consistently fills these varied geometries with realistic SAR backscatter, the downstream segmentation model is forced to reduce its reliance on shape heuristics and instead prioritize local radiometric features (e.g., dampening contrast, texture gradients) to identify the class. This explains the significant reduction in Look-alike false positives: the model learns that "dark texture" is a more reliable predictor than elongated shape. Finally, the smaller deltas on Swin-based models echo reports that transformer backbones encode strong global priors and multi-scale context~\cite{Cao2021SwinUnetUPA}, leaving less headroom for gains unless synthetic realism and weighting are tuned with care.

\paragraph{Methodological Insights}

\begin{itemize}
\item \textbf{The Synthetic Scale "Sweet Spot":} Contrary to the assumption that "more data is better," we observed diminishing returns when quadrupling the synthetic pool to $N{=}3608$. While Oil IoU remained high, Ship detection degraded. This suggests that simply repeating the stochastic generation process without increasing the diversity of the underlying seed patches leads to redundancy. A pool size of $N{=}1804$ (roughly $2\times$ the available real training seeds) provided the optimal balance between diversity and class distribution.
\item \textbf{Architecture-Specific Sensitivity:} The contrast between the ResNet-DeepLabV3+ (sensitive to augmentation) and Swin-Tiny UNet (robust/insensitive) highlights a fundamental difference in inductive biases. The Swin Transformer's hierarchical attention windows capture long-range dependencies natively~\cite{Cao2021SwinUnetUPA}, rendering it less dependent on synthetic context expansion. For CNNs, however, MORP-Synth acts as a critical "context regularizer," bringing their performance on par with Transformers for minority classes.
\item \textbf{Training Dynamics:} The use of a composite Focal Tversky loss was essential. By penalizing False Negatives and specifically weighting the confusion between Oil and Look-alikes, the loss function allowed the optimizer to exploit the subtle boundary cues provided by the synthetic samples, rather than collapsing into the majority "Sea" class. See Appendix ~\ref{tab:baseline_wce_dice}.
\end{itemize}

\paragraph{Operational Implications and Methodological Recommendations}
For Peruvian enforcement agencies such as OEFA, the reduction of False Positives (look-alikes) is the metric of highest strategic value. Given the prevalence of biogenic look-alikes in the Humboldt Current, a low False Positive Rate is essential to prevent "alert fatigue" and ensure interdiction resources are not wasted on false alarms ~\cite{bianchiLargeScaleDetectionCategorization2020,al-ruzouqSensorsFeaturesMachine2020}. The practical viability of this precision-oriented approach is validated by the model's temporal stability on major historical events. In the \textbf{Lobitos sequence (2024–2025)}, the model successfully tracked the slick's expansion from $\approx 5$~km$^2$ to $86$~km$^2$ over 15 days. This temporal coherence tracking a dispersing plume over 15 days without latching onto transient wind features confirms that the learned features are physically robust rather than artifacts of single-frame overfitting. Similarly, in the \textbf{Repsol event}, the model correctly identified the northward drift consistent with hydrodynamic forcing, despite severe nearshore clutter. Methodologically, these findings advocate for a \textit{backbone-aware} augmentation policy. We observe that controlled synthetic diversity substantially aids mid-capacity CNNs (acting as a context regularizer), whereas Transformer backbones benefit from conservative synthetic exposure. Broadly, this establishes the Peruvian SAR dataset with labeling aligned to Mediterranean source as a concrete testbed for quantifying transfer gaps in realistic coastal conditions.

\paragraph{Limitations and Future directions}
While MORP-Synth provides a data-efficient pathway for adaptation, several limitations remain. First, the morphological perturbations in MORP are geometric heuristics (apex discovery and displacement) rather than hydrodynamically constrained mask generation, without oceanographic priors they can miss plausible geometry from SAR appearance, especially for small objects. Second, while INADE delivers useful realism for supervision, it remains below recent diffusion-based syntheses in fine-scale texture diversity~\cite{moonDiffusionbasedDataAugmentation2025,wangSemanticImageSynthesis2022,kuangSemanticLayoutGuidedImageSynthesis2023}; our choice trades ultimate fidelity for efficiency and larger spatial contexts~\cite{tan2021inade}. Finally, the physical detection limits of C-band SAR impose an inherent ceiling on performance. The False Negatives present in the dataset likely reflects thin-sheen annotation uncertainty under low wind speeds, suggesting that future work should incorporate boundary-aware metrics (e.g., contour F-score) or multi-modal data to better quantify perceptual accuracy in these marginal regimes.

\section{Conclusions}
\label{sec:conclusions}

This work addressed the challenge of cross-domain oil spill segmentation by adapting deep models from Mediterranean source data to the Peruvian coast using the proposed \textbf{MORP-Synth} augmentation. Our results confirm that while standard fine-tuning bridges the initial domain gap, geometrically diversified synthetic data is essential for resolving the severe class imbalance typical of coastal monitoring.

The MORP-Synth framework improved the ResNet-34 DeepLabV3\texttt{+} baseline by 6.00 percentage points in mean IoU on the unseen test set. Most critically, it achieved a +14.59 pp gain in the discrimination of Look-alikes, validating our hypothesis that decoupling geometric priors from radiometric texture forces the model to learn more robust, physically grounded features. We observed an architecture dependent response: mid-capacity CNNs benefited substantially from synthetic regularization (acting as a context regularizer), whereas Transformer-based backbones (Swin-UNet) exhibited intrinsic robustness to spatial variations, leaving less headroom for synthetic gains.

In summary, MORP-Synth offers a data-efficient pathway to enhance SAR oil-spill detection in data scarce regions. By generating diverse, label consistent training samples, it reduces the dependency on extensive local annotation, reinforcing both the scientific understanding of marine pattern recognition and practical environmental response capacity.

\section*{Data and Code Availability}
Sentinel-1 SAR imagery is publicly available from the Copernicus Open Access Hub.

The Peruvian ground-truth masks and metadata cannot be will be uploaded to Zenodo 
upon manuscript acceptance.

All code for MORP--Synth (INADE training, MORP perturbations, and segmentation 
training) is available at: 

\texttt{https://github.com/andrexandrex/MorpSynth}

\onecolumn

\bibliographystyle{IEEEtran}
\bibliography{references}

\begin{thebibliography}{10}
\providecommand{\url}[1]{#1}
\csname url@samestyle\endcsname
\providecommand{\newblock}{\relax}
\providecommand{\bibinfo}[2]{#2}
\providecommand{\BIBentrySTDinterwordspacing}{\spaceskip=0pt\relax}
\providecommand{\BIBentryALTinterwordstretchfactor}{4}
\providecommand{\BIBentryALTinterwordspacing}{\spaceskip=\fontdimen2\font plus
\BIBentryALTinterwordstretchfactor\fontdimen3\font minus \fontdimen4\font\relax}
\providecommand{\BIBforeignlanguage}[2]{{%
\expandafter\ifx\csname l@#1\endcsname\relax
\typeout{** WARNING: IEEEtran.bst: No hyphenation pattern has been}%
\typeout{** loaded for the language `#1'. Using the pattern for}%
\typeout{** the default language instead.}%
\else
\language=\csname l@#1\endcsname
\fi
#2}}
\providecommand{\BIBdecl}{\relax}
\BIBdecl

\bibitem{bianchiLargeScaleDetectionCategorization2020}
F.~M. Bianchi, M.~M. Espeseth, N.~Borch, F.~M. Bianchi, M.~M. Espeseth, and N.~Borch, ``Large-{{Scale Detection}} and {{Categorization}} of {{Oil Spills}} from {{SAR Images}} with {{Deep Learning}},'' \emph{Remote Sensing}, vol.~12, no.~14, Jul. 2020.

\bibitem{al-ruzouqSensorsFeaturesMachine2020}
R.~{Al-Ruzouq}, M.~B.~A. Gibril, A.~Shanableh, A.~Kais, O.~Hamed, S.~{Al-Mansoori}, and M.~A. Khalil, ``Sensors, {{Features}}, and {{Machine Learning}} for {{Oil Spill Detection}} and {{Monitoring}}: {{A Review}},'' \emph{Remote Sensing}, vol.~12, no.~20, p. 3338, Oct. 2020.

\bibitem{garcia-pinedaClassificationOilSpill2020}
O.~{Garcia-Pineda}, G.~Staples, C.~E. Jones, C.~Hu, B.~Holt, V.~Kourafalou, G.~Graettinger, L.~DiPinto, E.~Ramirez, D.~Streett, J.~Cho, G.~A. Swayze, S.~Sun, D.~Garcia, and F.~{Haces-Garcia}, ``Classification of oil spill by thicknesses using multiple remote sensors,'' https://repository.library.noaa.gov, 2020.

\bibitem{wanStudyVariationPatterns2024}
Y.~Wan, X.~Chen, L.~Peng, H.~Wang, and R.~Zhang, ``Study on the variation patterns of sea surface oil spill characteristics based on {{GNSS-R}} under different wind speeds,'' \emph{Marine Pollution Bulletin}, vol. 208, p. 117005, Nov. 2024.

\bibitem{krestenitisOilSpillIdentification2019}
\BIBentryALTinterwordspacing
M.~Krestenitis, G.~Orfanidis, K.~Ioannidis, K.~Avgerinakis, S.~Vrochidis, and I.~Kompatsiaris, ``Oil {{Spill Identification}} from {{Satellite Images Using Deep Neural Networks}},'' vol.~11, no.~15, p. 1762. [Online]. Available: \url{https://www.mdpi.com/2072-4292/11/15/1762}
\BIBentrySTDinterwordspacing

\bibitem{krestenitisEarlyIdentificationOil2019}
------, ``Early {{Identification}} of {{Oil Spills}} in {{Satellite Images Using Deep CNNs}},'' in \emph{{{MultiMedia Modeling}}}, I.~Kompatsiaris, B.~Huet, V.~Mezaris, C.~Gurrin, W.-H. Cheng, and S.~Vrochidis, Eds.\hskip 1em plus 0.5em minus 0.4em\relax Cham: Springer International Publishing, 2019, vol. 11295, pp. 424--435.

\bibitem{ronnebergerUNetConvolutionalNetworks2015}
\BIBentryALTinterwordspacing
O.~Ronneberger, P.~Fischer, and T.~Brox, ``U-{{Net}}: {{Convolutional Networks}} for {{Biomedical Image Segmentation}}.'' [Online]. Available: \url{http://arxiv.org/abs/1505.04597}
\BIBentrySTDinterwordspacing

\bibitem{chenDeepLabSemanticImage2017}
\BIBentryALTinterwordspacing
L.-C. Chen, G.~Papandreou, I.~Kokkinos, K.~Murphy, and A.~L. Yuille, ``{{DeepLab}}: {{Semantic Image Segmentation}} with {{Deep Convolutional Nets}}, {{Atrous Convolution}}, and {{Fully Connected CRFs}}.'' [Online]. Available: \url{http://arxiv.org/abs/1606.00915}
\BIBentrySTDinterwordspacing

\bibitem{zhuOilSpillContextual2022}
\BIBentryALTinterwordspacing
Q.~Zhu, Y.~Zhang, Z.~Li, X.~Yan, Q.~Guan, Y.~Zhong, L.~Zhang, and D.~Li, ``Oil {{Spill Contextual}} and {{Boundary-Supervised Detection Network Based}} on {{Marine SAR Images}},'' vol.~60, pp. 1--10. [Online]. Available: \url{https://ieeexplore.ieee.org/document/9568691/}
\BIBentrySTDinterwordspacing

\bibitem{satyanarayanaOilSpillSegmentation2023}
\BIBentryALTinterwordspacing
A.~R. Satyanarayana and M.~A. Dhali, ``Oil {{Spill Segmentation}} using {{Deep Encoder-Decoder}} models.'' [Online]. Available: \url{http://arxiv.org/abs/2305.01386}
\BIBentrySTDinterwordspacing

\bibitem{dehghani-dehcheshmehOilSpillsDetection2023}
\BIBentryALTinterwordspacing
S.~Dehghani-Dehcheshmeh, M.~Akhoondzadeh, and S.~Homayouni, ``Oil spills detection from {{SAR Earth}} observations based on a hybrid {{CNN}} transformer networks,'' vol. 190, p. 114834. [Online]. Available: \url{https://linkinghub.elsevier.com/retrieve/pii/S0025326X23002655}
\BIBentrySTDinterwordspacing

\bibitem{buiOilSpillDetection2024}
\BIBentryALTinterwordspacing
N.~A. Bui, Y.~Oh, and I.~Lee, ``Oil spill detection and classification through deep learning and tailored data augmentation,'' vol. 129, p. 103845. [Online]. Available: \url{https://linkinghub.elsevier.com/retrieve/pii/S1569843224001997}
\BIBentrySTDinterwordspacing

\bibitem{cuiEnhancedUnsupervisedDomain2025}
G.~Cui, J.~Fan, and Y.~Zou, ``Enhanced unsupervised domain adaptation with iterative pseudo-label refinement for inter-event oil spill segmentation in {{SAR}} images,'' \emph{International Journal of Applied Earth Observation and Geoinformation}, vol. 139, p. 104479, May 2025.

\bibitem{kussulTransferLearningSinglepolarized2025}
N.~Kussul, Y.~Salii, V.~Kuzin, B.~Yailymov, and A.~Shelestov, ``Transfer learning and single-polarized {{SAR}} image preprocessing for oil spill detection,'' \emph{ISPRS Open Journal of Photogrammetry and Remote Sensing}, vol.~15, p. 100081, Jan. 2025.

\bibitem{mogollonREPSOLOilSpill2023}
\BIBentryALTinterwordspacing
R.~Mogollón, C.~Arellano, P.~Villegas, D.~Espinoza-Morriberón, and J.~Tam, ``{{REPSOL}} oil spill off {{Central Perú}} in {{January}} 2022: {{A}} modeling case study,'' vol. 194, p. 115282. [Online]. Available: \url{https://linkinghub.elsevier.com/retrieve/pii/S0025326X23007166}
\BIBentrySTDinterwordspacing

\bibitem{Wang2018DeepDA}
M.~Wang and W.~Deng, ``Deep visual domain adaptation: A survey,'' \emph{Neurocomputing}, vol. 312, pp. 135--153, 2018.

\bibitem{pratapFineArtFinetuning2025}
\BIBentryALTinterwordspacing
S.~Pratap, A.~R. Aranha, D.~Kumar, G.~Malhotra, A.~P.~N. Iyer, and S.~S.s., ``The fine art of fine-tuning: {{A}} structured review of advanced {{LLM}} fine-tuning techniques,'' vol.~11, p. 100144. [Online]. Available: \url{https://www.sciencedirect.com/science/article/pii/S2949719125000202}
\BIBentrySTDinterwordspacing

\bibitem{moonDiffusionbasedDataAugmentation2025}
\BIBentryALTinterwordspacing
J.~Moon, J.~Yun, J.~Kim, J.~Lee, and M.~Kim, ``Diffusion-based {{Data Augmentation}} and {{Knowledge Distillation}} with {{Generated Soft Labels Solving Data Scarcity Problems}} of {{SAR Oil Spill Segmentation}},'' \emph{arXiv preprint arXiv:2412.08116}. [Online]. Available: \url{http://arxiv.org/abs/2412.08116}
\BIBentrySTDinterwordspacing

\bibitem{sunUtilizingDeepLearning2024}
Z.~Sun, Q.~Yang, N.~Yan, S.~Chen, J.~Zhu, J.~Zhao, and S.~Sun, ``Utilizing deep learning algorithms for automated oil spill detection in medium resolution optical imagery,'' \emph{Marine Pollution Bulletin}, vol. 206, p. 116777, Sep. 2024.

\bibitem{buiIMPROVINGACCURACYOIL2023}
\BIBentryALTinterwordspacing
N.~A. Bui, Y.~G. Oh, and I.~P. Lee, ``{{IMPROVING THE ACCURACY OF AN OIL SPILL DETECTION AND CLASSIFICATION MODEL WITH FAKE DATASETS}},'' vol. X-1-W1-2023, pp. 51--56. [Online]. Available: \url{https://isprs-annals.copernicus.org/articles/X-1-W1-2023/51/2023/isprs-annals-X-1-W1-2023-51-2023.html}
\BIBentrySTDinterwordspacing

\bibitem{Park2019SPADE}
T.~Park, M.-Y. Liu, T.-C. Wang, and J.-Y. Zhu, ``Semantic image synthesis with spatially-adaptive normalization,'' in \emph{CVPR}, 2019, pp. 2337--2346.

\bibitem{tan2021inade}
Z.~Tan, D.~Chen, Q.~Chu, M.~Chai, J.~Liao, M.~He, L.~Yuan, G.~Hua, and N.~Yu, ``Efficient semantic image synthesis via class-adaptive normalization,'' \emph{IEEE Transactions on Pattern Analysis and Machine Intelligence}, 2021.

\bibitem{Abraham2019FocalTversky}
N.~Abraham and N.~Khan, ``A novel focal tversky loss function with improved attention u-net for lesion segmentation,'' in \emph{2019 IEEE 16th International Symposium on Biomedical Imaging (ISBI)}, 2019, pp. 683--687.

\bibitem{Salehi2017Tversky}
S.~S.~M. Salehi, D.~Erdogan, and A.~Gholipour, ``Tversky loss function for image segmentation using 3d fully convolutional deep networks,'' in \emph{Medical Image Computing and Computer Assisted Intervention – MLMI}, 2017, pp. 379--387.

\bibitem{abrahamNovelFocalTversky2018}
N.~Abraham and N.~M. Khan, ``A {{Novel Focal Tversky}} loss function with improved {{Attention U-Net}} for lesion segmentation,'' October 2018.

\bibitem{cuiClassBalancedLossBased2019}
Y.~Cui, M.~Jia, T.-Y. Lin, Y.~Song, and S.~Belongie, ``Class-{{Balanced Loss Based}} on {{Effective Number}} of {{Samples}},'' January 2019.

\bibitem{leeSpeckleAnalysisSmoothing1981}
J.-S. Lee, ``Speckle analysis and smoothing of synthetic aperture radar images,'' \emph{Computer Graphics and Image Processing}, vol.~17, no.~1, pp. 24--32, Sep. 1981.

\bibitem{ronnebergerUNet2015}
O.~Ronneberger, P.~Fischer, and T.~Brox, ``U-net: Convolutional networks for biomedical image segmentation,'' in \emph{MICCAI}, 2015.

\bibitem{heDeepResidual2015}
K.~He, X.~Zhang, S.~Ren, and J.~Sun, ``Deep residual learning for image recognition,'' in \emph{CVPR}, 2016.

\bibitem{chenDeepLab2017}
L.-C. Chen, G.~Papandreou, F.~Schroff, and H.~Adam, ``Rethinking atrous convolution for semantic image segmentation,'' in \emph{arXiv}, 2017.

\bibitem{tan2021efficientnetv2}
M.~Tan and Q.~Le, ``Efficientnetv2: Smaller models and faster training,'' in \emph{Proceedings of the 38th International Conference on Machine Learning}, 2021, pp. 10\,096--10\,106.

\bibitem{chen2017rethinking}
L.-C. Chen, G.~Papandreou, F.~Schroff, and H.~Adam, ``Rethinking atrous convolution for semantic image segmentation,'' \emph{arXiv preprint arXiv:1706.05587}, 2017.

\bibitem{Cao2021SwinUnetUPA}
\BIBentryALTinterwordspacing
H.~Cao, Y.~Wang, J.~Chen, D.~Jiang, X.~Zhang, Q.~Tian, and M.~Wang, ``Swin-unet: Unet-like pure transformer for medical image segmentation,'' in \emph{MICCAI Workshop}, 2021. [Online]. Available: \url{https://arxiv.org/abs/2105.05537}
\BIBentrySTDinterwordspacing

\bibitem{Chen2021TransUNet}
\BIBentryALTinterwordspacing
J.~Chen, Y.~Xie, J.~Wang, Z.~Zhang \emph{et~al.}, ``Transunet: Transformers make strong encoders for medical image segmentation,'' in \emph{MICCAI}, 2021. [Online]. Available: \url{https://arxiv.org/abs/2102.04306}
\BIBentrySTDinterwordspacing

\bibitem{savitzkySmoothingDifferentiationData1964}
{\relax Abraham}.~Savitzky and M.~J.~E. Golay, ``Smoothing and {{Differentiation}} of {{Data}} by {{Simplified Least Squares Procedures}}.'' \emph{Analytical Chemistry}, vol.~36, no.~8, pp. 1627--1639, Jul. 1964.

\bibitem{parkSemanticImageSynthesis2019}
\BIBentryALTinterwordspacing
T.~Park, M.-Y. Liu, T.-C. Wang, and J.-Y. Zhu, ``Semantic {{Image Synthesis}} with {{Spatially-Adaptive Normalization}}.'' [Online]. Available: \url{http://arxiv.org/abs/1903.07291}
\BIBentrySTDinterwordspacing

\bibitem{dhariwalDiffusionModelsBeat2021}
P.~Dhariwal and A.~Nichol, ``Diffusion {{Models Beat GANs}} on {{Image Synthesis}},'' in \emph{Advances in {{Neural Information Processing Systems}}}, vol.~34.\hskip 1em plus 0.5em minus 0.4em\relax Curran Associates, Inc., 2021, pp. 8780--8794.

\bibitem{yangDiffusionModelsComprehensive2025}
L.~Yang, Z.~Zhang, Y.~Song, S.~Hong, R.~Xu, Y.~Zhao, W.~Zhang, B.~Cui, and M.-H. Yang, ``Diffusion {{Models}}: {{A Comprehensive Survey}} of {{Methods}} and {{Applications}},'' Sep. 2025.

\bibitem{heWeightingMethodsRare2021}
\BIBentryALTinterwordspacing
J.~He and M.~X. Cheng, ``Weighting {{Methods}} for {{Rare Event Identification From Imbalanced Datasets}},'' vol.~4. [Online]. Available: \url{https://www.frontiersin.org/journals/big-data/articles/10.3389/fdata.2021.715320/full}
\BIBentrySTDinterwordspacing

\bibitem{salehiTverskyLossFunction2017}
S.~S.~M. Salehi, D.~Erdogmus, and A.~Gholipour, ``Tversky loss function for image segmentation using {{3D}} fully convolutional deep networks,'' June 2017.

\bibitem{liHardNegativesMining2025}
\BIBentryALTinterwordspacing
G.~Li, Y.~Gao, X.~Huang, B.~W.-K. Ling, G.~Li, Y.~Gao, X.~Huang, and B.~W.-K. Ling, ``A {{Hard Negatives Mining}} and {{Enhancing Method}} for {{Multi-Modal Contrastive Learning}},'' vol.~14, no.~4. [Online]. Available: \url{https://www.mdpi.com/2079-9292/14/4/767}
\BIBentrySTDinterwordspacing

\bibitem{tanDiverseSemanticImage2021}
\BIBentryALTinterwordspacing
Z.~Tan, M.~Chai, D.~Chen, J.~Liao, Q.~Chu, B.~Liu, G.~Hua, and N.~Yu, ``Diverse {{Semantic Image Synthesis}} via {{Probability Distribution Modeling}},'' in \emph{2021 {{IEEE}}/{{CVF Conference}} on {{Computer Vision}} and {{Pattern Recognition}} ({{CVPR}})}.\hskip 1em plus 0.5em minus 0.4em\relax IEEE, pp. 7958--7967. [Online]. Available: \url{https://ieeexplore.ieee.org/document/9577617/}
\BIBentrySTDinterwordspacing

\bibitem{heuselGANsTrainedTwo2018}
M.~Heusel, H.~Ramsauer, T.~Unterthiner, B.~Nessler, and S.~Hochreiter, ``{{GANs Trained}} by a {{Two Time-Scale Update Rule Converge}} to a {{Local Nash Equilibrium}},'' January 2018.

\bibitem{luoLearningSmoothHinge2021}
J.~Luo, H.~Qiao, and B.~Zhang, ``Learning with {{Smooth Hinge Losses}},'' March 2021.

\bibitem{chenTransUNetTransformersMake2021}
\BIBentryALTinterwordspacing
J.~Chen, Y.~Lu, Q.~Yu, X.~Luo, E.~Adeli, Y.~Wang, L.~Lu, A.~L. Yuille, and Y.~Zhou, ``{{TransUNet}}: {{Transformers Make Strong Encoders}} for {{Medical Image Segmentation}}.'' [Online]. Available: \url{http://arxiv.org/abs/2102.04306}
\BIBentrySTDinterwordspacing

\bibitem{charngDeepLearningSegmentation2020}
J.~Charng, D.~Xiao, M.~Mehdizadeh, M.~Attia, S.~Arunachalam, T.~Lamey, J.~Thompson, T.~McLaren, J.~Roach, D.~Mackey, S.~Frost, and F.~Chen, ``Deep learning segmentation of hyperautofluorescent fleck lesions in {{Stargardt}} disease,'' vol.~10, p. 16491.

\bibitem{wangSinGANLabelerEnhancedSinGAN2025}
B.~Wang, L.~Chen, D.~Song, W.~Chen, J.~Yu, B.~Wang, L.~Chen, D.~Song, W.~Chen, and J.~Yu, ``{{SinGAN-Labeler}}: {{An Enhanced SinGAN}} for {{Generating Marine Oil Spill SAR Images}} with {{Labels}},'' \emph{Journal of Marine Science and Engineering}, vol.~13, no.~3, Feb. 2025.

\bibitem{wangSemanticImageSynthesis2022}
\BIBentryALTinterwordspacing
W.~Wang, J.~Bao, W.~Zhou, D.~Chen, D.~Chen, L.~Yuan, and H.~Li, ``Semantic {{Image Synthesis}} via {{Diffusion Models}},'' \emph{arXiv preprint arXiv:2207.00050}. [Online]. Available: \url{http://arxiv.org/abs/2207.00050}
\BIBentrySTDinterwordspacing

\bibitem{kuangSemanticLayoutGuidedImageSynthesis2023}
\BIBentryALTinterwordspacing
Y.~Kuang, F.~Ma, F.~Li, Y.~Liu, and F.~Zhang, ``Semantic-{{Layout-Guided Image Synthesis}} for {{High-Quality Synthetic-Aperature Radar Detection Sample Generation}},'' vol.~15, no.~24, p. 5654. [Online]. Available: \url{https://www.mdpi.com/2072-4292/15/24/5654}
\BIBentrySTDinterwordspacing

\end{thebibliography}

\section{Appendix}

\appendix

\begin{table}[p]
\centering
\caption{Sentinel-1 SAR images used in this study, ordered by spill date. Includes acquisition delay and patch counts for training and testing subsets.}
\label{tab:peruvian_selection}
\scriptsize
\begin{tabular}{|c|c|c|c|c|c|c|c|}
\hline
\textbf{Img} & \textbf{Spill Date} & \textbf{Latitude} & \textbf{Longitude} & \textbf{Acquisition Date} & \textbf{$\Delta$ Days} & \textbf{Patches} & \textbf{Split} \\
\hline
1 & 22/09/2014 & -80.7519 & -3.5860 & 11/10/2014 & 19 & 38 & Train \\
2 & 14/03/2015 & -81.2819 & -4.5736 & 16/03/2015 & 2 & 64 & Train \\
3 & 05/01/2016 & -71.3392 & -17.6346 & 13/01/2016 & 8 & 52 & Train \\
4 & 11/03/2016 & -79.0288 & -8.1117 & 26/03/2016 & 15 & 88 & Train \\
5 & 19/05/2016 & -81.2745 & -4.5963 & 30/05/2016 & 11 & 18 & Train \\
6 & 08/08/2016 & -71.3393 & -17.6346 & 22/08/2016 & 14 & 157 & Train \\
7 & 27/03/2017 & -81.2922 & -4.4562 & 13/04/2017 & 17 & 68 & Train \\
8 & 27/03/2017 & -81.2922 & -4.4562 & 01/04/2017 & 5 & 56 & Train \\
9 & 27/03/2017 & -81.2922 & -4.4562 & 16/04/2017 & 20 & 44 & Train \\
10 & 11/07/2017 & -81.2717 & -4.3194 & 18/07/2017 & 7 & 6 & Train \\
11 & 11/07/2017 & -81.2717 & -4.3194 & 21/07/2017 & 10 & 32 & Train \\
12 & 05/06/2018 & -77.1448 & -12.0472 & 17/06/2018 & 12 & 157 & Train \\
13 & 11/10/2018 & -77.1457 & -12.0472 & 27/10/2018 & 16 & 70 & Train \\
14 & 19/11/2018 & -79.0288 & -8.1117 & 29/11/2018 & 10 & 58 & Test \\
15 & 07/04/2019 & -81.3123 & -4.6213 & 18/04/2019 & 11 & 10 & Train \\
16 & 18/04/2019 & -81.2604 & -4.2810 & 18/04/2019 & 0 & 24 & Train \\
17 & 18/04/2019 & -81.2604 & -4.2810 & 30/04/2019 & 12 & 60 & Test \\
18 & 04/09/2019 & -81.3656 & -4.4692 & 09/09/2019 & 5 & 22 & Test \\
19 & 14/10/2019 & -81.2777 & -4.3096 & 27/10/2019 & 13 & 32 & Test \\
20 & 18/03/2020 & -81.2349 & -4.2436 & 31/03/2020 & 13 & 74 & Train \\
21 & 18/03/2020 & -81.2349 & -4.2436 & 19/03/2020 & 1 & 22 & Test \\
22 & 17/08/2020 & -81.3086 & -4.6100 & 22/08/2020 & 5 & 32 & Train \\
23 & 09/01/2021 & -81.3235 & -4.7120 & 25/01/2021 & 16 & 38 & Train \\
24 & 28/04/2021 & -81.2725 & -4.2896 & 10/05/2021 & 12 & 26 & Train \\
25 & 17/05/2021 & -81.3299 & -4.7144 & 22/05/2021 & 5 & 62 & Test \\
26 & 17/05/2021 & -81.3299 & -4.7144 & 25/05/2021 & 8 & 26 & Test \\
27 & 02/11/2021 & -81.3048 & -4.2936 & 18/11/2021 & 16 & 18 & Train \\
28 & 04/11/2021 & -81.2940 & -4.4414 & 18/11/2021 & 14 & 30 & Test \\
29 & 15/01/2022 & -77.1520 & -11.9256 & 02/02/2022 & 18 & 76 & Train \\
30 & 15/01/2022 & -77.1520 & -11.9256 & 25/01/2022 & 10 & 158 & Train \\
31 & 15/01/2022 & -77.1680 & -11.9638 & 25/01/2022 & 10 & 160 & Train \\
32 & 24/08/2022 & -81.2725 & -4.2896 & 30/08/2022 & 6 & 14 & Train \\
33 & 09/10/2022 & -81.0984 & -4.5111 & 17/10/2022 & 8 & 66 & Train \\
34 & 18/08/2023 & -81.2921 & -4.3025 & 25/08/2023 & 7 & 54 & Train \\
35 & 18/08/2023 & -81.2921 & -4.3025 & 03/09/2023 & 16 & 30 & Train \\
36 & 18/08/2023 & -81.2921 & -4.3025 & 22/08/2023 & 4 & 38 & Train \\
37 & 21/12/2024 & -81.3319 & -4.5322 & 07/01/2025 & 17 & 58 & Train \\
38 & 21/12/2024 & -81.3319 & -4.5322 & 29/12/2024 & 8 & 40 & Train \\
39 & 21/12/2024 & -81.3319 & -4.5322 & 26/12/2024 & 5 & 20 & Train \\
40 & 21/12/2024 & -81.3319 & -4.5322 & 10/01/2025 & 20 & 34 & Train \\
\hline
\end{tabular}
\end{table}
\clearpage

\begin{algorithm}[H]
\caption{MORP (Morphological Region Perturbation)}
\label{alg:morp}
\begin{algorithmic}[1]
\Require Label map $\mathcal{L}\in\{0,\dots,4\}^{H\times W}$; target classes $\mathcal{C}{=}\{1,2\}$
\Require Selection policy: \textsc{SelectRegions}$(\cdot;\,k,\mathrm{mode})$, diversity flag
\Require Placement params: angle range $[-\pi,\pi)$; max shift $S_{\max}$; ...
\Require Apex discovery params: $(w,p,q,d,\rho,d_s)$
\Require Apex selection params: $m$ (apices per region)
\Require Apex edit params: $(\alpha, n_{\mathrm{rays}})$, $(s_{\mathrm{exp}}, \dots), (r_{\max}^{\mathrm{exp}}, \dots), p_{\mathrm{exp}}$
\Statex \Comment{\textit{Note: Placement, discovery, and edit params may be class-specific.}}
\Ensure Augmented labels $\mathcal{L}^\star$

\State $\mathcal{L}^\star \gets \mathcal{L}$
\State $\mathcal{R} \gets \bigcup_{c\in\mathcal{C}}\mathrm{ConnComp}(\mathds{1}[\mathcal{L}{=}c])$
\State $\{R_i\}*{i=1}^K \gets \textsc{SelectRegions}(\mathcal{R};\,k{=}n*{\mathrm{reg}},\mathrm{mode},\mathrm{diversity})$
\Comment{class-balanced if requested}

\Statex \hfill\textit{// Stage 1: Flat-Aware Region Placement}
\For{$i=1$ \textbf{to} $K$}
\State $c_i \gets \mathrm{Class}(R_i)$
\State sample $\theta\sim\mathcal{U}[-\pi,\pi)$, $\Delta$ with $\|\Delta\|*2\le S*{\max}$
\State $R^{\mathrm{rot}}*i \gets \mathrm{RotateNoCrop}(R_i,\theta)$
\State $R^{\mathrm{fa}}i \gets \textsc{FlatAwareBulges}(R^{\mathrm{rot}}i;\,\alpha,n{\mathrm{rays}},s{\mathrm{exp}},r*{\max}^{\mathrm{exp}})$
\If{$\mathrm{TryPaste}(\mathcal{L}^\star,R^{\mathrm{fa}}*i,\Delta;\ \Omega*{\text{forbid}},\Omega_{\text{allow}})$}
\State \textbf{continue}
\Else
\State $\mathrm{RestoreNearOrigin}(\mathcal{L}^\star,R^{\mathrm{fa}}_i,\text{center-align})$
\EndIf
\EndFor

\Statex \hfill\textit{// Stage 2 \& 3: Apex Detection and Perturbation}
\State $\mathcal{R}' \gets \bigcup_{c\in\mathcal{C}}\mathrm{ConnComp}(\mathds{1}[\mathcal{L}^\star{=}c])$
\State $\{Q_j\}*{j=1}^M \gets \textsc{SelectRegions}(\mathcal{R}';\,k{=}K,\mathrm{mode})$
\State $Q*{\mathrm{large}} \gets$ largest oil in $\mathcal{R}'$ if $|Q|/(HW)\ge\gamma$; ensure $Q_{\mathrm{large}}\in\{Q_j\}$

\For{$j=1$ \textbf{to} $M$}
\State $R'\gets Q_j$, $c\gets\mathrm{Class}(R')$, $A\gets\emptyset$
\State $A \gets \textsc{ApexDetectionViaSmoothedCurvature}(R';\,w,p,q,d,\rho,d_s)$
\hfill\Comment{See Algorithm \ref{alg:apices}}
\If{$|A|=0$} \textbf{continue} \EndIf
\State $A_m \gets \textsc{SelectApicesByKMeans}(A;\,m)$ \hfill\Comment{one per cluster, farthest from centroid}
\State $(B,S) \gets \textsc{ApexEditMulti}\Big(
R',A_m;\ \alpha,n_{\mathrm{rays}},\ p_{\mathrm{exp}}(c),\ (s_{\mathrm{exp}},s_{\mathrm{shr}})(c),\ (r_{\max}^{\mathrm{exp}},r_{\max}^{\mathrm{shr}})(c),\ Q_{\mathrm{large}}\Big)$
\hfill\Comment{See Algorithm \ref{alg:apexedit}}
\State $\mathcal{L}^\star \gets \mathrm{Write}\big((R'\cup B)\setminus S \text{ at label }c\big)$
\State $\mathcal{L}^\star \gets \textsc{RemoveSmall}(\mathcal{L}^\star;\,\text{min\_px})$
\EndFor
\State \Return $\mathcal{L}^\star$
\end{algorithmic}
\end{algorithm}

\begin{algorithm}[H]
\caption{Apex Detection via Smoothed Curvature}
\label{alg:apices}
\begin{algorithmic}[1]
\Require Region mask $R'$, smoothing window $w$, polyorder $p$, peak prominence quantile $q$, radial boost $\rho$, step $d_s$, min dist $d$
\Ensure Set of apex coordinates $A$
\State $P \gets \mathrm{OuterContour}(R')$ \Comment{ordered $(x_t,y_t)$, $t{=}1..N$}
\State $(\tilde{x},\tilde{y}) \gets \mathrm{SavitzkyGolay}(x,y; w,p,\text{mode=wrap})$
\State compute discrete curvature $\kappa_t$ from $(\tilde{x}_t,\tilde{y}_t)$ using step $d_s$
\State $\kappa^{+}_t \gets \mathrm{RadialBoost}(\max(0,\kappa_t), \rho)$
\State $\tau \gets \mathrm{Quantile}(\{\kappa^{+}_t\},q)$
\State $\text{peaks} \gets \mathrm{FindPeaks}(\kappa^{+},\text{prominence}>\tau,\text{distance}\ge d)$
\State \Return $A \gets \{(x_t,y_t)\,:\, t\in\text{peaks}\}$
\end{algorithmic}
\end{algorithm}

\begin{algorithm}[H]
\caption{Apex-based Perturbation (Single Apex Edit)}
\label{alg:apexedit}
\begin{algorithmic}[1]
\Require Region mask $R'$, apex $a$, fan half-angle $\alpha$, ray count $n_{\mathrm{rays}}$
\Require Prob.\ $p_{\mathrm{exp}}$; scales $(s_{\mathrm{exp}}, s_{\mathrm{shr}})$; max radii $(r_{\max}^{\mathrm{exp}}, r_{\max}^{\mathrm{shr}})$
\Ensure Add-mask $B$ (bulge), remove-mask $S$ (wedge)
\State $B,S \gets \emptyset,\emptyset$
\State $\vec{n} \gets \mathrm{NormalAt}(R',a)$ \Comment{e.g., gradient of distance transform}
\State $z \sim \mathrm{Bernoulli}(p_{\mathrm{exp}})$
\If{$z{=}1$} \Comment{expand}
\State $B \gets \mathrm{RadialFanPolygon}(a,\vec{n},\alpha,n_{\mathrm{rays}}, s_{\mathrm{exp}}, r_{\max}^{\mathrm{exp}},\text{...})$
\Else \Comment{shrink}
\State $S \gets \mathrm{RadialFanPolygon}(a,-\vec{n},\alpha,n_{\mathrm{rays}}, s_{\mathrm{shr}}, r_{\max}^{\mathrm{shr}},\text{...}) \cap R'$
\EndIf
\State \Return $B,S$
\end{algorithmic}
\end{algorithm}

\paragraph{Symbols of MORP algorithm.}
\begin{center}
\begin{tabular}{@{}ll@{}}
\toprule
Symbol & Meaning \\
\midrule
$\mathcal{L}, \mathcal{L}^\star$ & Input and augmented label maps \\
$\mathcal{C}$ & Target classes $\{1,2\}$ (\emph{oil}, \emph{look-alike}) \\
$\mathcal{R}, \mathcal{R}'$ & Set of all connected components (before/after placement) \\
$R, Q$ & A single connected component (region) \\
$K, M$ & Number of selected regions for placement / editing \\
$A, A_m$ & Set of all apices; $m$-sized subset of chosen apices \\
$B, S$ & Add-mask (bulge), remove-mask (shrink) \\
$H, W$ & Height and Width of the label map \\
$\gamma$ & Area threshold to detect a "large" oil region \\
$c$ & Class label (e.g., 1 or 2) \\
$S_{\max}$ & Max shift distance for placement (Stage 1) \\
$w, p$ & Savitzky-Golay filter window size and polynomial order \\
$q, d, \rho, d_s$ & Apex discovery params (quantile, distance, boost, step) \\
$m$ & Number of apices to select per region \\
$\alpha, n_{\mathrm{rays}}$ & Fan half-angle and ray count for edits \\
$r_{\max}^{\mathrm{exp}}, r_{\max}^{\mathrm{shr}}$ & Max radial extent for expand/shrink (Stage 3) \\
$s_{\mathrm{exp}}, s_{\mathrm{shr}}$ & Scale factors for expand/shrink (Stage 3) \\
$p_{\mathrm{exp}}$ & Probability of performing an "expand" operation \\
\bottomrule
\end{tabular}
\end{center}

\begin{table}[htbp]
\centering
\caption{Baseline fine-tuned segmentation performance on the Peruvian dataset using the standard Weighted Cross Entropy + Dice loss.}
\label{tab:baseline_wce_dice}
\resizebox{\textwidth}{!}{
\begin{tabular}{lcccccccc} 
\toprule
\textbf{Model} & \textbf{Epoch (ckpt)} & \textbf{Pixel Acc.} & \textbf{mIoU} & \textbf{Sea (0)} & \textbf{Oil (1)} & \textbf{Look-alike (2)} & \textbf{Ship (3)} & \textbf{Land (4)} \\
\midrule
ResNet-34 DeepLabV3\texttt{+} & 88 & 0.9345 & 0.5085 & 0.9471 & 0.3085 & 0.1582 & 0.2190 & 0.9097 \\
ResNet-34 UNet+ASPP           & 86 & 0.9189 & 0.5190 & 0.9240 & 0.3777 & 0.1257 & 0.2844 & 0.8831 \\
ResNet-34 UNet                & 71 & 0.9453 & 0.5255 & 0.9481 & 0.4505 & 0.1087 & 0.2308 & 0.8893 \\
\bottomrule
\end{tabular}
}
\end{table}

\begin{table}[htbp]
\centering
\caption{Effect of synthetic \texttt{morph} (mask growth) and \texttt{scale} (synthetic dataset size) with multi-scale augmentation during mixed training (\textbf{Swin-Unet-Tiny}, real $N = 902$). Values are IoU (\%).}
\label{tab:swin_unet_tiny_synth_morph_scale_multi}
\small
\resizebox{\textwidth}{!}{%
\begin{tabular}{c
ccc
ccc
ccc
ccc
ccc
ccc
}
\toprule
\textbf{Scale}
& \multicolumn{3}{c}{\textbf{mIoU}}
& \multicolumn{3}{c}{\textbf{Sea}}
& \multicolumn{3}{c}{\textbf{Oil}}
& \multicolumn{3}{c}{\textbf{Look-alike}}
& \multicolumn{3}{c}{\textbf{Ship}}
& \multicolumn{3}{c}{\textbf{Land}} \\
\cmidrule(lr){2-4}\cmidrule(lr){5-7}\cmidrule(lr){8-10}\cmidrule(lr){11-13}\cmidrule(lr){14-16}\cmidrule(lr){17-19}
& m00 & m50 & m100 & m00 & m50 & m100 & m00 & m50 & m100 & m00 & m50 & m100 & m00 & m50 & m100 & m00 & m50 & m100 \\
\midrule
25  & 54.73 & 55.05 & 60.23 & 94.71 & 93.35 & 95.07 & 40.31 & 46.12 & 51.25 & 25.79 & 25.56 & 37.20 & 21.61 & 19.73 & 26.07 & 91.22 & 90.49 & 91.55 \\
50  & 57.38 & 57.00 & 57.46 & 95.57 & 93.47 & 93.90 & 43.66 & 47.18 & 51.06 & 26.57 & 28.86 & 30.29 & 27.36 & 23.74 & 20.00 & 93.75 & 91.65 & 92.06 \\
100 & 53.65 & 57.70 & 56.70 & 92.99 & 94.03 & 92.92 & 41.99 & 48.91 & 50.60 & 23.50 & 31.17 & 28.07 & 18.63 & 23.21 & 21.04 & 91.12 & 91.19 & 90.84 \\
150 & 53.47 & 56.35 & 56.80 & 92.66 & 93.79 & 93.60 & 51.88 & 50.42 & 48.41 & 21.92 & 28.20 & 28.44 & 11.26 & 16.61 & 21.11 & 89.65 & 90.55 & 92.23 \\
200 & 57.44 & 58.58 & 57.39 & 95.05 & 93.72 & 94.66 & 50.95 & 51.39 & 51.14 & 30.21 & 30.64 & 30.64 & 21.06 & 25.49 & 20.13 & 89.94 & 91.65 & 90.24 \\
\bottomrule
\end{tabular}}
\end{table}

\begin{figure}[htbp]
\centering

\begin{tabular*}{\textwidth}{@{\extracolsep{\fill}} c c c c c @{}}
\scriptsize \textbf{(a) Original} &
\scriptsize \textbf{(b) Rotate \& Place + apex} &
\scriptsize \textbf{(c) Curvature (+/-)} &
\scriptsize \textbf{(d) Final image} &
\scriptsize \textbf{(e) Synth SAR} \\
\end{tabular*}

\vspace{0.5em}

\includegraphics[width=\textwidth]{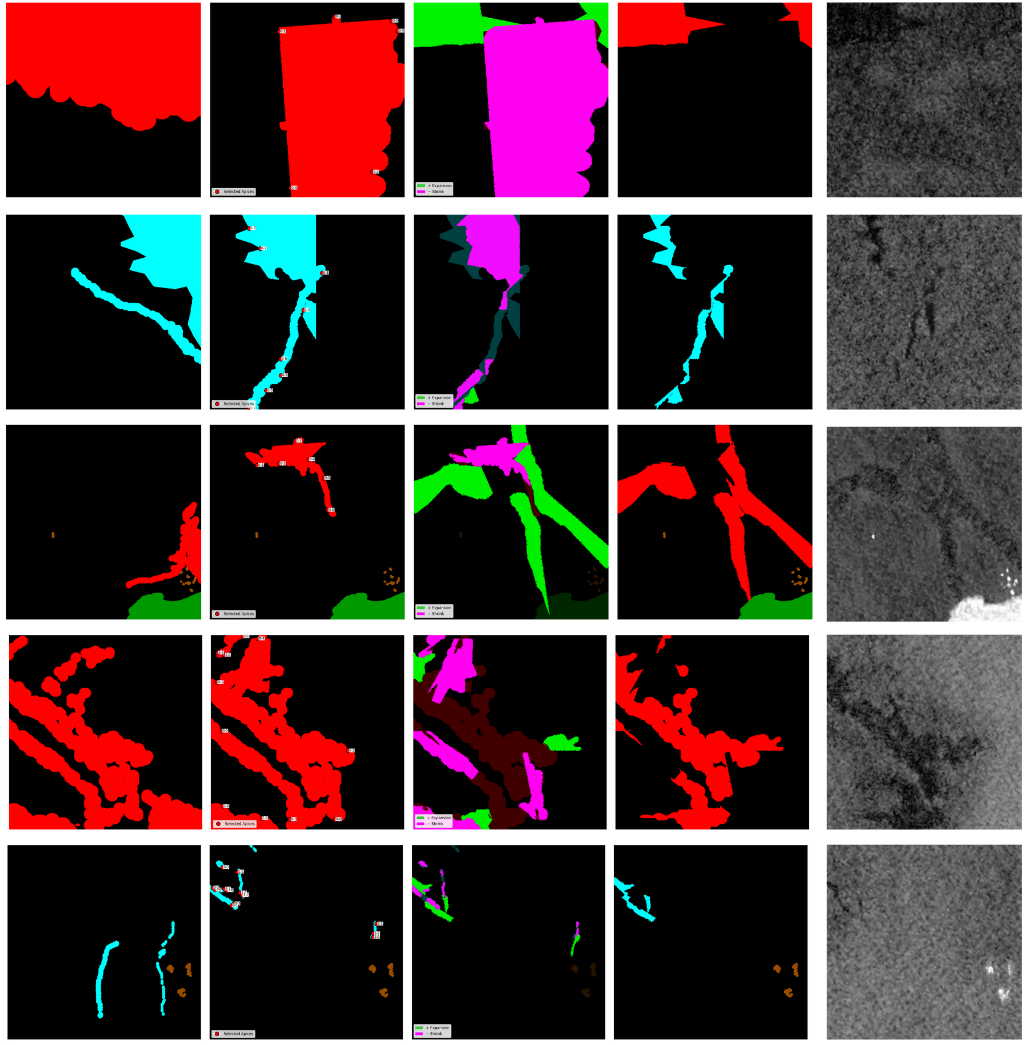}

\caption{Illustration of the complete augmentation workflow including the Synth SAR (INADE) output. Each column represents:
(a) original image,
(b) rotated and positioned mask with apex selection,
(c) curvature-based perturbations indicating expansions and shrinkages,
(d) final augmented result, and
(e) synthesized SAR patch.}
\label{fig:morp_workflow_unified}
\end{figure}

\end{document}